\documentclass{article}

\usepackage{graphicx}
\usepackage{amsmath}
\usepackage{amssymb}
\usepackage{mathtools}
\usepackage{amsthm}
\usepackage[table]{xcolor}
\usepackage{subcaption}
\usepackage{wrapfig}
\usepackage{microtype}
\usepackage{booktabs}
\usepackage{threeparttable}
\usepackage{multirow}  
\usepackage[utf8]{inputenc} 
\usepackage[sorting=none,doi=false,isbn=false,url=false]{biblatex}
\theoremstyle{remark}
\newtheorem{remark}{Remark}
\usepackage{enumitem}

\usepackage[final,nonatbib]{neurips_2025}

\usepackage[T1]{fontenc}    
\usepackage[colorlinks=true, urlcolor=black]{hyperref}
\usepackage{url}            
\usepackage{booktabs}       
\usepackage{amsfonts}       
\usepackage{nicefrac}       
\usepackage{microtype}      
\usepackage{xcolor}         

\title{One Filters All: A Generalist Filter \\ for State Estimation}

\author{
  Shiqi Liu\textsuperscript{1 *},  \
  Wenhan Cao\textsuperscript{1 *},
 Chang Liu\textsuperscript{2}, 
  Zeyu He\textsuperscript{1},
  Tianyi Zhang\textsuperscript{1},
  Shengbo Eben Li\textsuperscript{1 3 \textdagger} \\ \\
  \textsuperscript{1} School of Vehicle and Mobility, Tsinghua University \\
    \textsuperscript{2} College of Engineering, Peking University \quad
    \textsuperscript{3} College of AI, Tsinghua University \\
    \textsuperscript{*}Equal contribution \quad \textsuperscript{\textdagger}Corresponding author
}

\begin{document}

\maketitle

\begin{abstract}
Estimating hidden states in dynamical systems, also known as optimal filtering, is a long-standing problem in various fields of science and engineering. 
In this paper,
we introduce a general filtering framework, 
, which leverages large language models (LLMs) for state estimation by embedding noisy observations with text prototypes. 
In various experiments for classical dynamical systems, we find that first, 
state estimation can significantly
benefit from the reasoning knowledge embedded in pre-trained LLMs. 
By achieving proper modality alignment
with the frozen LLM,
LLM-Filter outperforms the state-of-the-art learning-based approaches.
Second, we carefully design the prompt structure, System-as-Prompt (SaP), incorporating task instructions that enable the LLM to understand the estimation tasks.
Guided by these prompts, LLM-Filter exhibits exceptional  generalization, capable of performing filtering tasks
accurately in changed or even unseen environments. 
We further observe
a scaling-law behavior in LLM-Filter, where accuracy improves with
larger model sizes and longer training times.
These findings make LLM-Filter a promising foundation model of filtering.
\end{abstract}

\section{Introduction}

State estimation of dynamical systems is a crucial topic in various fields, including robotics \parencite{barfoot2024state}, meteorology \parencite{course2023state}, chemistry \parencite{dochain2003state}, and transportation \parencite{zhang2024physics}.
The most comprehensive framework for state estimation is Bayesian filtering \parencite{sarkka2023bayesian}, which performs online estimation by iteratively applying prediction and update steps. Popular online Bayes filters can be categorized into Gaussian filters \parencite{ito2000gaussian} and particle filters (PFs) \parencite{chen2003bayesian}.
In high-dimensional non-Gaussian systems, Gaussian filters tend to produce significant errors \parencite{afshari2017gaussian, straka2021importance}, whereas PFs are limited by their substantial computational demands \parencite{ van2019particle,revach2022kalmannet}.

\begin{figure}[t]
\begin{center}
\centerline{\includegraphics[width=0.95\columnwidth]{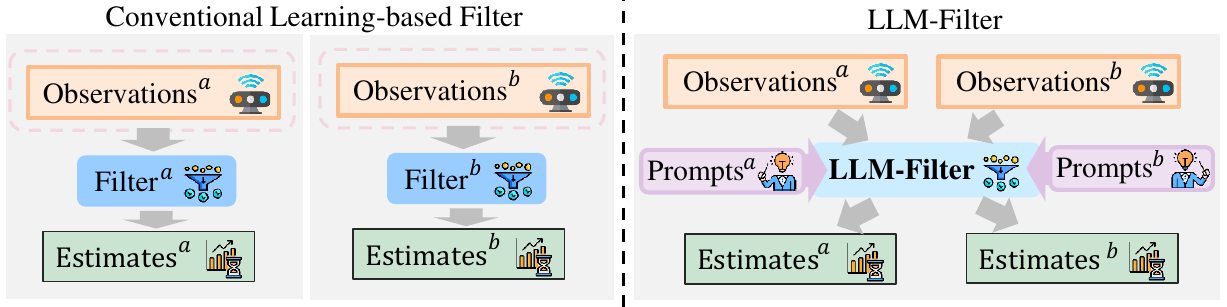}}
\caption{\textbf{Comparisons of LLM-Filter with Conventional Learning based Filters.}
The left subfigure shows conventional learning-based filters, which rely on specific system data and training. The right subfigure shows LLM-Filter, which generalizes across different systems through prompt guidance, without requiring retraining.
}
\label{fig.motivation}
\vskip -.3in
\end{center}
\end{figure}

To address these limitations, there is a growing trend toward learning-based filtering methods \parencite{tang2021reinforcement, revach2022kalmannet, ji2022concurrent, dahal2024robuststatenet}. Trained offline on well-curated datasets, these learning-based filters directly capture accurate system statistics rather than relying on manual modeling, enabling high accuracy and efficient online estimation. However, these filters are typically tailored to specific tasks and require retraining when applied to different dynamical systems, as illustrated in Figure~\ref{fig.motivation}. Without retraining, their performance often deteriorates when the system undergoes changes or transitions to entirely new environments.

The control domain faces analogous challenges when adapting to diverse tasks and environments \parencite{ahn2022can,team2024octo}.
Large control models have been proposed to address these challenges, with examples like RT \parencite{brohan2022rt, brohan2023rt} and OpenVLA \parencite{kim2024openvla}.  
By leveraging the large-scale pretraining knowledge of large language models (LLMs) and vision-language models (VLMs), these large control models demonstrate effective generalization across diverse robots, tasks, and environments \parencite{o2023open}.
It is worth noting that, \emph{control is intrinsically a dual problem of filtering}, first demonstrated in linear systems through Lagrangian duality \parencite{kailath2000linear, goodwin2005lagrangian}, and later extended to nonlinear systems \parencite{todorov2008general, kim2022duality}.
In the filtering domain, to the best of the authors' knowledge, no filtering method has yet integrated LLMs to utilize pre-training knowledge, nor has a general filtering model been developed.

To fill this research gap, we propose LLM-Filter, a generalist filter that reprograms LLMs for state estimation while preserving the integrity of the backbone model. 
The process begins by embedding noisy observations as text prototypes, which are subsequently processed by a frozen LLM to harmonize disparate data modalities.
To enable generalization cross systems, we augment inputs with \emph{System-as-Prompt} (SaP), which includes task-specific instructions and task examples.
The outputs generated by the LLM are then mapped onto the final state estimates.
By leveraging the pretraining knowledge of the LLM and guiding it through carefully designed prompts, LLM-Filter can outperform most state-of-the-art learning-based filters and exhibit remarkable in-context generalization, even when applied to completely new systems, as demonstrated in our experiments. 
Our main contributions are as follows:

\begin{itemize}
\item We demonstrate that state estimation can significantly benefit from the knowledge embedded in pre-trained language models. 
By establishing proper modality alignment between the filtering process and the inference mechanism of LLMs, we propose a generalist filter, called LLM-Filter.
\item 
We carefully design an in-context approach, SaP, to help the LLM understand the current task and adapt to the specific application system. Guided by SaP, LLM-Filter exhibits exceptional generalization, effectively performing filtering tasks in unseen scenarios.
\item 
In our experiments, LLM-Filter consistently outperforms the state-of-the-art learning-based filters and demonstrates remarkable in-context generalization in changed or entirely new systems without retraining or fine-tuning.
Additionally, we observe a scaling-law behavior, where accuracy improves with larger model sizes and longer training times.

\end{itemize}

\section{Preliminaries}
\subsection{Formulation of State Estimation}
\label{sub.Formulation of State Estimation}
We begin with a brief review of state estimation and the primary filtering methods considered in this work. Typically, state estimation is modeled using a Markovian state-space model:
\begin{subequations} \label{eq.ssm}
\begin{align}
\boldsymbol{x}_{t+1}&=f\left(\boldsymbol{x}_{t}\right)+\boldsymbol{\xi}_{t}, \label{eq.ssm_transiton}\\
\boldsymbol{y}_{t}&=h\left(\boldsymbol{x}_{t}\right)+\boldsymbol{\zeta}_{t},\label{eq.ssm_obs}
\end{align}
\end{subequations}
for $t\in\{1,2,\dots\}$. Here $f(\cdot)$ and $h(\cdot)$ represent the state transition and observation models, respectively. The term $\boldsymbol{x}_t\in  \mathbb{R}^M$ is defined as the state while $\boldsymbol{y}_t\in \mathbb{R}^N$ denotes the observations. 
Furthermore, $\boldsymbol{\xi}_t$ and  $\boldsymbol{\zeta}_t$ represent the transition noise and observation noise, respectively, with their probability distributions generally assumed to be known.
This state-space model description in \eqref{eq.ssm} can be represented as a hidden Markov model: 
\begin{equation}\label{eq.hmm}
\begin{aligned}
\boldsymbol{x}_{0} &\sim p(\boldsymbol{x}_0),
\\
\boldsymbol{x}_{t} &\sim p(\boldsymbol{x}_t|\boldsymbol{x}_{t-1}),
\\
\boldsymbol{y}_{t} &\sim p(\boldsymbol{y}_t|\boldsymbol{x}_{t}),
\end{aligned}  
\end{equation}
where, $p(\boldsymbol{x}_t|\boldsymbol{x}_{t-1})$ and $p(\boldsymbol{y}_{t}|\boldsymbol{x}_{t})$ are the transition and output probabilities, respectively, and $p(\boldsymbol{x}_0)$ denotes the initial state distribution.
Essentially, both \eqref{eq.ssm} and \eqref{eq.hmm} provide different formulations of the same underlying system model.
The objective of state estimation is to extract the information of the state $\boldsymbol{x}_t$ on all available observations ${\boldsymbol{y}_{1:t}}$.

\subsection{Online Bayesian Filtering} 
A principled framework for state estimation is Bayesian filtering, which computes $p(\boldsymbol{x}_t|\boldsymbol{y}_{1:t})$ recursively through two steps:
\begin{subequations}\label{eq.BF}
\begin{align}
p(\boldsymbol{x}_t|\boldsymbol{y}_{1:t-1}) &= \int p(\boldsymbol{x}_t|\boldsymbol{x}_{t-1}) p(\boldsymbol{x}_{t-1}|\boldsymbol{y}_{1:t-1}) \mathrm{d}  \boldsymbol{x}_{t-1}, \label{eq.prediction}
\\
p(\boldsymbol{x}_t|\boldsymbol{y}_{1:t}) &= \frac{p(\boldsymbol{y}_t|\boldsymbol{x}_t) p(\boldsymbol{x}_t|\boldsymbol{y}_{1:t-1})}{\int p(\boldsymbol{y}_t|\boldsymbol{x}_t) p(\boldsymbol{x}_t|\boldsymbol{y}_{1:t-1}) \mathrm{d} \boldsymbol{x}_t },
\label{eq.update}
\end{align}   
\end{subequations}
where, \eqref{eq.prediction} is called 
prediction step while \eqref{eq.update} is called the update step.
Under the assumptions of linearity and Gaussian noise, \eqref{eq.BF} can be analytically calculated online, which is the basis of Kalman filter (KF) \parencite{kalman1960new}, as detailed in Appendix \ref{apd.filter_methods}.
However, these filtering methods are highly dependent on accurate modeling $f(\cdot)$ and $h(\cdot)$, as well as the noise distributions of $\boldsymbol{\xi}_t$ and $\boldsymbol{\zeta}_t$ specific to each system.

\subsection{Learning-based Filtering}\label{sub.learing-based-filter}
Instead of performing online computations as in \eqref{eq.BF}, learning-based filtering methods are trained on well-collected data to learn the correlation between observations and the underlying state. This learning is achieved by optimizing the following supervised learning loss to update the network parameters $\boldsymbol{\theta}$:
\begin{equation}\label{eq.learning-based_loss}
\mathcal{L}(\boldsymbol{\theta}) = ||{\boldsymbol{x}}_t-\hat{\boldsymbol{x}}_t(\boldsymbol{y}_{1:t},\boldsymbol{\theta})||_2^2,
\end{equation}
where $\hat{\boldsymbol{x}}_t(\boldsymbol{y}_{1:t},\boldsymbol{\theta})$
represents the estimates, and $\boldsymbol{x}_t$ is the ground truth.
By capturing accurate system statistics from training data instead of relying on manual modeling, learning-based filtering methods often provide more precise estimates and greater efficiency. 
However, the heavy reliance on task-specific datasets renders conventional learning-based filters inflexible, requiring retraining whenever tasks change or transition to entirely new environments.
To overcome this limitation,  this paper aims to propose a general filtering method that leverages an in-context mechanism, enabling effective performance across multiple systems without retraining.

\section{Methodology}
\begin{figure*}[ht]
\begin{center}
\centerline{\includegraphics[width=\textwidth]{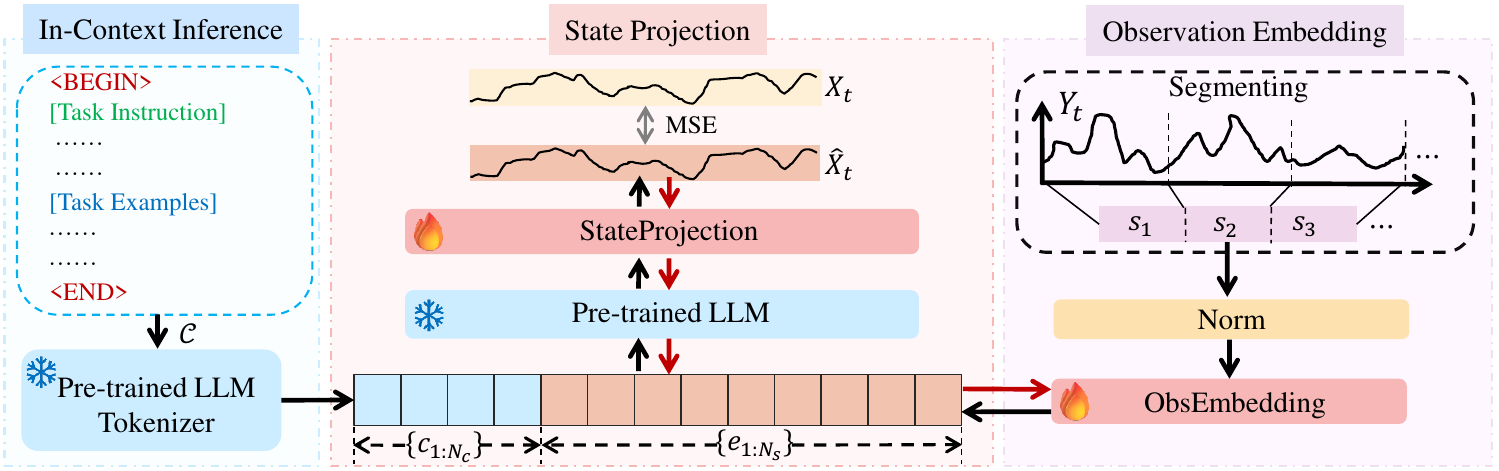}}
\caption{\textbf{Framework of LLM-Filter}. (1) \textbf{Observation Embedding:} Observations are segmented and embedded into token embeddings.
(2) \textbf{In-context Inference:} In-context tokens from the tokenized SaP context are concatenated with the observation embeddings.
(3) \textbf{State Projection:} The LLM's predicted tokens are projected into the state space to obtain the final estimates.
}
\label{fig.framework_3}
\vskip -.3in
\end{center}
\end{figure*}

The main structure of LLM-Filter is illustrated in Figure~\ref{fig.framework_3}.
We first align the modalities between the discrete tokens of the LLM and continuous observations (Section.\ref{sub.modality_alignment}),
then design an in-context inference approach based on system information to enhance generalization (Section.\ref{sub.in-context}),  
finally obtain the estimates from the predicted tokens of the LLM and define the loss function (Section.\ref{sub.token_process}).

\subsection{Observation Embedding}
\label{sub.modality_alignment}
\textbf{Moving Horizon.}
As described in Section.\ref{sub.Formulation of State Estimation}, the objective is to estimate the state $\boldsymbol{x}_t$ on all available observations ${\boldsymbol{y}_{1:t}}$.
However, it is impractical to input an ever-growing sequence of observations. Inspired by the moving horizon estimation \parencite{alessandri2008moving} and the context window of LLMs \parencite{achiam2023gpt}
, we design a sliding window with a fixed length $T$ for estimation.
The input observations are defined as $\boldsymbol{Y}_t=\{\boldsymbol{y}_{t-T+1:t}\} \in \mathbb{R}^{T \times N}$ , and the corresponding output estimates are $\boldsymbol{\hat{X}}_t=\{\boldsymbol{\hat{x}}_{t:t+T-1}\} \in \mathbb{R}^{T \times M}$ with $\boldsymbol{{X}}_t=\{\boldsymbol{{x}}_{t:t+T-1}\} \in \mathbb{R}^{T \times M}$ representing the ground truth states, where $N$ and $M$ denote the dimensions of the observation and state variables, respectively.

\textbf{Segment and Embedding.}
To align the continuous observations with the discrete tokens, 
we segment the input observations and embed them into LLMs.
A popular segmentation method uses a \textit{single-series sequence} format \parencite{liu2024autotimes, liutimer}, which involves directly flattening all data dimensions. However, this approach can \textbf{disrupt the inherent correlations between variables}, such as the crucial relationship between position and speed, which is vital for accurate position estimation.
To preserve the inherent correlations, we embed $\boldsymbol{Y}_t$ by segmenting it based on segment length $L$ while preserving its multi-dimensional structure:
\begin{equation}
    \nonumber
    \boldsymbol{Y}_t = \{\boldsymbol{s}_{1}, \boldsymbol{s}_{2}, \cdots, \boldsymbol{s}_{N_s} \}, \quad N_s = \left\lfloor \frac{T}{L} \right\rfloor,
\end{equation}
where each segment $\boldsymbol{s}_{i} \in \mathbb{R}^{L \times N}$ (for $i = 1, 2, \dots, N_s$) and $\left\lfloor \cdot \right\rfloor$ denotes the floor function. If the division is not exact, padding is applied at the end to ensure all segments have the same length $L$.
To leverage the pretraining knowledge and the intrinsic token transitions of the LLM, we freeze the parameters and eliminate the embedding and projection layers designed for language tokens. Instead, we introduce $\operatorname{ObsEmbedding(\cdot)}: \mathbb{R}^{L \times N} \mapsto \mathbb{R}^{D \times N}$, which independently embeds each normalized segment into the $D$-dimensional latent space of the LLM:
\begin{equation}
    \nonumber
    \boldsymbol{e}_i = \operatorname{ObsEmbedding}(\operatorname{Norm}(\boldsymbol{s}_{i})), \quad i = 1, 2, \dots, N_s.
\end{equation}

\begin{wrapfigure}{r}{0.5\textwidth}
\vspace{-10pt}
\centering \includegraphics[width=0.5\textwidth]
{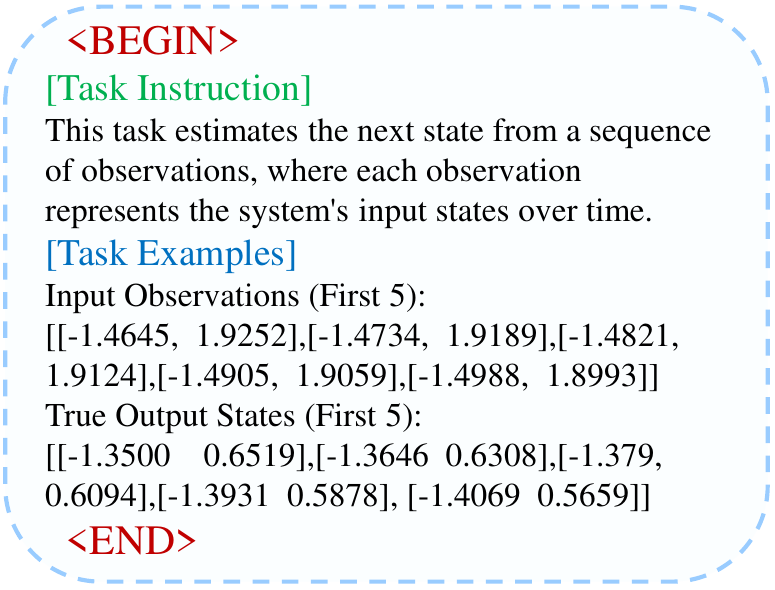}
\caption{An Illustration of SaP for Hopf \parencite{hassard1981theory} system .}
\label{fig.prompt}
\end{wrapfigure}

\subsection{In-context Inference}
\label{sub.in-context}
Traditional learning-based filters estimate the true system states solely from observational data, as discussed in Section~\ref{sub.learing-based-filter}. However, this reliance limits their ability to identify underlying system dynamics, resulting in poor generalization to new systems.
To address this, we build upon the in-context learning capabilities of LLMs \parencite{yin2023survey} and introduce a novel prompting strategy called SaP, which guides  LLM-Filter to adapt flexibly to different systems.
SaP is grounded in two components necessary for crafting effective prompts for LLM-Filter: Task Instruction and Task Examples.
\textbf{Task Instruction} equips LLM-Filter with crucial contextual knowledge, which may differ across various domains.
\textbf{Task Examples} guide LLM-Filter by showcasing example scenarios that clarify the task at hand.
An example depiction of SaP can be found in Figure~\ref{fig.prompt}.
When inference, the SaP text $\mathcal{C}$ will be fed to the pre-trained LLM tokenizer for processing language text:
\begin{equation}
    \nonumber
    \left\{\boldsymbol{c}_1, \cdots, \boldsymbol{c}_{N_c}\right\} =
    \operatorname{LLMTokenizer}(\mathcal{C}),
\end{equation}
where $N_c$ is the number of tokenized context tokens. 

\subsection{State Projection}
\textbf{Token Prediction.}
Prevalent LLMs, trained on diverse downstream tasks, excel at predicting the next token based on previous ones across various domains \parencite{driess2023palm, han2024onellm}. By inputting the guiding SaP and the embedding token of a previous observation, we aim to derive state estimation features for the next period.
To achieve modality alignment, we utilize only the core layers of the LLM by excluding the original embedding and projection layers for language tokens. 
The tokenized SaP context and observation embeddings $\left\{\boldsymbol{c}_1, \cdots, \boldsymbol{c}_{N_c},\boldsymbol{e}_1, \cdots, \boldsymbol{e}_{N_s}\right\}$ are directly fed  into the core layers, which then generate the corresponding output token embeddings $\left\{\hat{\boldsymbol{e}}_1, \cdots, \hat{\boldsymbol{e}}_{N_{s}}\right\}$:
\begin{equation}
\begin{aligned}
\label{eq.token_prediction}
 \hat{E}=& \operatorname{LLMCoreLayers}\left(\left\{\boldsymbol{c}_1, \cdots, \boldsymbol{c}_{N_c},\boldsymbol{e}_1, \cdots, \boldsymbol{e}_{N_s}\right\}\right), \\
 &\left\{\hat{\boldsymbol{e}}_1, \cdots, \hat{\boldsymbol{e}}_{N_{s}}\right\} = \hat{E}_{\{N_c+1:N_c+N_s\}}.
\end{aligned}
\end{equation}

\textbf{Output Projection.}
Afterward, we employ the layer $\operatorname{StateProjection}(\cdot): \mathbb{R}^{D \times N} \mapsto \mathbb{R}^{T \times M}$ to project the output embeddings $\left\{\hat{\boldsymbol{e}}_1, \cdots, \hat{\boldsymbol{e}}_{N_{s}}\right\}$ onto the system state space, thereby obtaining the final estimates $\hat{\boldsymbol{X}}_t$:
\begin{equation}
    \nonumber
\hat{\boldsymbol{X}}_t=\operatorname{StateProjection}(\left\{\hat{\boldsymbol{e}}_1, \cdots, \hat{\boldsymbol{e}}_{N_{s}}\right\}),
\end{equation}
where both $\operatorname{ObsEmbedding(\cdot)}$ and $\operatorname{StateProjection(\cdot)}$ are implemented using simple MLPs, which has been shown to effectively embed and project time series for LLM  \parencite{liu2024autotimes}.



\textbf{Loss Function.}
\label{sub.token_process}
Finally, the overall objective is to minimize the error between the ground truth ${\boldsymbol{X}}_t$ and the output estimations $\hat{\boldsymbol{X}}_t$ to optimize the parameters $\boldsymbol{\theta}$ of LLM-Filter $\mathcal{F}_{\boldsymbol{\theta}}$:
\begin{equation}\nonumber
\mathcal{L}_{\boldsymbol{\theta}} = \frac{1}{T}||{\boldsymbol{X}}_t-\hat{\boldsymbol{X}}_t||_2^2
=\frac{1}{T}||{\boldsymbol{X}}_t-
\mathcal{F}_{\boldsymbol{\theta}}(\boldsymbol{Y}_{t},\mathcal{C})||_2^2.
\end{equation}
This loss function can be seen as an extension of \eqref{eq.learning-based_loss}, incorporating additional context input.
Notably, to preserve the pre-trained knowledge and due to resource constraints, we freeze the LLM and only update the parameters of $\operatorname{ObsEmbedding(\cdot)}$ and $\operatorname{StateProjection}$.
The detailed configuration of LLM-Filter is provided in Appendix \ref{apd.implementation}.
\begin{remark}
When the SaP prompt is not available, LLM-Filter can still operate without in-context guidance, a variant we refer to as \textbf{LLM-Filter-O}. However, due to the lack of contextual guidance, its filtering accuracy and generalization capability may degrade, as demonstrated in the comparison provided in Section~\ref{sec.experiments}.
\end{remark}

\section{Related Work}
\textbf{Online Bayesian Filtering.}
The field of state estimation can be traced back to the pioneering development of the KF \parencite{kalman1960new}, which was famously applied to NASA's Apollo guidance and navigation systems at the Ames Research Center \parencite{grewal2010applications}.  This milestone led to advanced methods like the extended Kalman filter (EKF) \parencite{smith1962application}, unscented Kalman filter (UKF) \parencite{julier1995new}, enabling applications in nonlinear systems. 
However, their reliance on linearization and Gaussian assumptions can introduce significant errors in complex scenarios. 
In response, particle-based filtering methods were developed, including the PF \parencite{liu1998sequential} and the ensemble Kalman filter (EnKF) \parencite{sakov2008implications}.
Unfortunately, particle-based approaches face high computational demands and the curse of dimensionality.

\textbf{Learning-based Filtering.}
Recent advances in neural network integration have driven the development of learning-based filtering methods. For instance, the KalmanNet \parencite{revach2022kalmannet, ni2024adaptive} adopts RNN to compute Kalman gains, significantly improving computational efficiency and accuracy. 
Similarly, MEstimator \parencite{ji2022concurrent} and RStateNet \parencite{dahal2024robuststatenet} utilize multilayer perceptrons (MLPs) and long short-term memory networks (LSTMs) \parencite{hochreiter1997long} to process historical state ensembles, enriching input information and achieving higher precision.
ProTran \parencite{tang2021probabilistic}, on the other hand, employs attention mechanisms to model system dynamics in the latent space, enabling long-term estimations. 
However, these methods rely solely on observational data as input and require retraining when applied to new systems.
In contrast, our proposed LLM-Filter aligns the modalities of language and state estimation, allowing it to generalize to unseen systems through in-context guidance without additional training.

\textbf{Large Control Model.}
As the dual problem of state estimation, there is an increasing trend toward training multi-task generalist control models on large, diverse datasets across different embodiments.
Recent work has incorporated LLMs and VLMs into control \parencite{ahn2022can, driess2023palm}, focusing primarily on high-level planning while underutilizing the knowledge from large-scale models.
To address this, the RT model \parencite{brohan2022rt,brohan2023rt,o2023open} directly train VLMs designed for open-vocabulary visual question answering to output low-level robot actions.
Building on this architecture, OpenVLA \parencite{kim2024openvla} streamlines the model's parameters while improving performance and exploring further fine-tuning strategies for generalization.
The rapid advancement of large control models drives our ambition to develop a generalist state estimation model that leverages pre-trained LLMs' knowledge.

\textbf{LLMs for Time Series Forecasting.}
State estimation and time series forecasting both use sequential data. However, state estimation estimates hidden states from noisy observations, while time series forecasting predicts future values from the same data.
Recent studies have made significant breakthroughs, especially in zero-shot forecasting, by establishing mappings between LLM tokens and numerical data \parencite{zhou2023one, gruver2024large}.
Additionally, TimeLLM \parencite{jin2023time}has explored the in-context forecasting capabilities of LLMs, leveraging an in-context learning to introduce a variety of forecasting tasks. AutoTimes \parencite{liu2024autotimes} demonstrates how timestamp encoding techniques can unlock further potential for LLMs in time-series applications. 
While these models primarily predict future values based on past data, LLM-Filter focuses on estimating hidden states from noisy observations in dynamic systems.

\section{Experiments} \label{sec.experiments}
The goal of our experimental evaluations is to  answer the following questions:
\begin{enumerate}
[itemsep=0pt,topsep=5pt,partopsep=-5pt,labelsep=0.5em,left=2pt]
    \item Can LLM-Filter learn to filter in canonical estimation task? (Section.\ref{subsec.canonical_estimation})
    \item Can LLM-Filter effectively generalize to new systems? (Section.\ref{subsec.zero-shot_estimation})
    \item Do the inherent characteristics of LLMs manifest within LLM-Filter? (Section.\ref{subsec.analysis_llm})
\end{enumerate}


\textbf{Benchmarks.} 
To comprehensively evaluate the performance of LLM-Filter, we conducted experiments on two types of classical dynamical systems: (1) low-dimensional nonlinear systems, and (2) high-dimensional chaotic systems, which are commonly used as benchmarks in previous studies \parencite{raanes2024dapper, duran2024outlier, course2023state}. Specifically, the low-dimensional nonlinear systems includes Selkov \parencite{sel1968self}, Oscillator \parencite{brunton2016discovering},  Hopf \parencite{hassard1981theory}, and Double Pendulum \parencite{levien1993double} systems.  For high-dimensional chaotic systems, we performed experiments on Lorenz96 \parencite{lorenz1996predictability} and VL20 \parencite{vissio2020mechanics}, both of which are widely used in atmospheric dynamics research. 
For the generalization experiments, we also include the Tracking system \parencite{sarkka2023bayesian}.
Detailed descriptions and configurations of each system are outlined in Appendix \ref{apd.system_description}.
We utilize root mean square error (RMSE) and average runtime (ms/step) as the primary evaluation metrics, with their precise definitions provided in Appendix \ref{apd.metrics}.

\textbf{Baselines.} We compare LLM-Filter with state-of-the-art methods, including online Bayes filters: the PF \parencite{van2019particle}, the EnKF \parencite{sakov2008implications}, the average-particle ensemble Kalman filter (EnKFI) \parencite{duran2024outlier}, the soft ensemble Kalman filter (EnKFS) \parencite{duran2024outlier}, and the huber ensemble Kalman filter (HubEnKF) \parencite{roh2013observation} as well as the learning-based methods: MEstimator \parencite{ji2022concurrent}, RStateNet \parencite{dahal2024robuststatenet}, ProTran \parencite{tang2021probabilistic}, and KalmanNet \parencite{revach2022kalmannet}. 
For a fair comparison, the hyperparameters of all online Bayes filters are selected during the first trial using the Bayesian Optimization (BO) package \parencite{gardner2014bayesian}.
Detailed descriptions of these methods are provided in Appendix \ref{apd.benchmar_settings}. 
Unless otherwise stated, LLaMA-7B \parencite{touvron2023llama} is used as the base LLM in LLM-Filter.

\subsection{Results for Canonical Estimation Task}
\label{subsec.canonical_estimation}
In this experiment, we evaluate the fundamental filtering capabilities of LLM-Filter and LLM-Filter-O on canonical estimation tasks, comparing their performance to state-of-the-art methods including MEstimator, RStateNet, ProTran, KalmanNet, EnKF, and PF. 


\textbf{Low-dimensional Nonlinear Systems.}
We first conducted experiments on low-dimensional nonlinear systems. A summary of the results is presented in Table~\ref{tab.nonlinear}, where LLM-Filter consistently outperform all other methods across all tested systems.
Notably, the RMSE of the LLM-Filter shows a maximum reduction of $32.00\%$ across all systems. On average, it demonstrates an improvement of $17.93\%$ over the best-performing learning-based filtering methods. When compared to the top-performing online Bayes filter, LLM-Filter still achieves an average improvement of $21.65\%$. 
In addition, with the use of the in-context prompt, LLM-Filter achieves a $10.18\%$ improvement over the data-only LLM-Filter-O, indicating that in-context prompting helps LLM-Filter better understand the system and perform more accurate state estimation.
These results highlight the strong filtering capabilities of the LLM-Filter in low-dimensional nonlinear systems, establishing its state-of-the-art performance.

\begin{table*}[ht]
\caption{RMSE of State Estimation on Classical Systems.
The best performances are marked in \textbf{bold}, and the second-best performances are \underline{underlined}.
Results labeled as \emph{NaN} indicate cases where the models diverged during training.}
\centering
\resizebox{\textwidth}{!}{
\begin{tabular}{lcccccccc}
\toprule
\multicolumn{1}{c}{\multirow{2}{*}{\textbf{Method}}} & \multicolumn{6}{c}{\textbf{Learning-Based Filters}} & \multicolumn{2}{c}{\textbf{Online Bayes Filters}} \\
\cmidrule(lr){2-7}  \cmidrule(lr){8-9}
& \cellcolor{gray!20}\textbf{LLM-Filter} & \cellcolor{gray!20}\textbf{LLM-Filter-O}& \textbf{MEstimator} & \textbf{RStateNet} & \textbf{ProTran} & \textbf{KalmanNet} & \textbf{EnKF} & \textbf{PF} \\
\midrule
\multicolumn{1}{c}{\textbf{Selkov}} & \cellcolor{gray!20}\textbf{0.4061} & \cellcolor{gray!20}0.6369 & 0.8864 & 0.7202 & 1.0219 & 1.1662 & \underline{0.5978} & 0.6863 \\
\multicolumn{1}{c}{\textbf{Oscillator}} & \cellcolor{gray!20}\textbf{0.5247} & \cellcolor{gray!20}0.5753 & 0.8347 & 0.8493 & 0.8933 & 0.5665 & \underline{0.5505} & 0.6807 \\
\multicolumn{1}{c}{\textbf{Hopf}} & \cellcolor{gray!20}\textbf{0.5751} & \cellcolor{gray!20}0.8180 & 0.8290 & 0.7282 & 0.7146 & 1.1984 & \underline{0.6322} & 0.6801 \\
\multicolumn{1}{c}{\textbf{Pendulum}} & \cellcolor{gray!20}\textbf{0.8348} & \cellcolor{gray!20}0.9218 & 0.9354 & 0.9180 & \underline{0.8456} & 2.7140 & 1.4117 & 5.3788 \\
\midrule
\multicolumn{1}{c}{\textbf{Lorenz96}} & \cellcolor{gray!20}\textbf{0.9149} & \cellcolor{gray!20}0.9735 & \underline{0.9649} & 0.9762 & 0.9975 & NaN & 6.6024 & 4.6289 \\
\multicolumn{1}{c}{\textbf{VL20}} & \cellcolor{gray!20}\textbf{0.7717} & \cellcolor{gray!20}0.8433 & 1.0014 & \underline{0.9428} & 0.9902 & NaN & 5.8633 & 11.6980 \\
\midrule
\multicolumn{1}{c}{\textbf{Average}} & \cellcolor{gray!20}\textbf{0.6712} & \cellcolor{gray!20}\underline{0.7948} & 0.9086 & 0.8558 & 0.9105 & NaN & 2.6097 & 3.9588 \\
\bottomrule
\end{tabular}}
\label{tab.nonlinear}
\end{table*}

\begin{figure}[ht]
\begin{center}
\centerline{\includegraphics[width=0.99\textwidth]{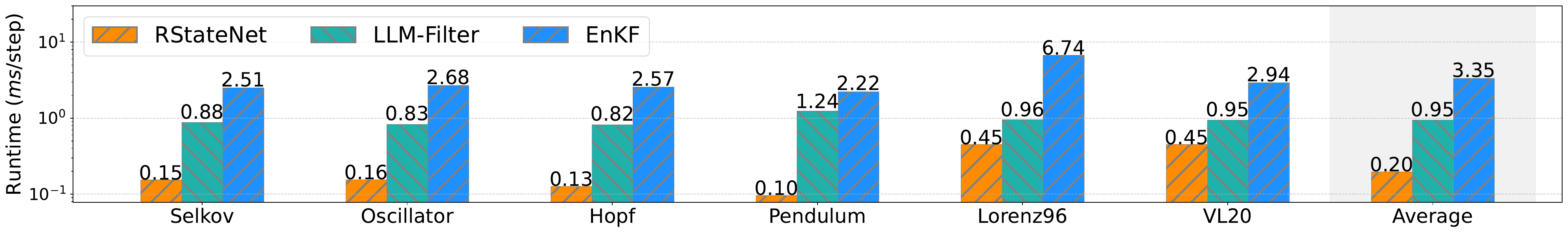}}
\caption{Comparison of Estimation Runtime Between LLM-Filter, EnKF, and RStateNet.
Comprehensive results are presented in Tab~\ref{tab.runtime}.}
\label{fig.runtime}
\end{center}
\vspace{-10pt}
\end{figure}

\textbf{\noindent{High-dimensional Chaotic Systems.}}
We then performed experiments on the Lorenz96 and VL20 systems, both with 72 state dimensions, designed to simulate atmospheric dynamics. The results are summarized in Table~\ref{tab.nonlinear}.
LLM-Filter achieves state-of-the-art performance on the Lorenz96 and VL20 systems, with an average improvement of $11.7\%$ over the best baseline filters. In contrast, the KalmanNet, as implemented in the official codebase\footnote{The official open-source repository is available at \href{https://github.com/KalmanNet/KalmanNet_TSP}{https://github.com/KalmanNet/KalmanNet\_TSP}. Despite optimizing the hyperparameters and network architecture, KalmanNet still struggles to filter high-dimensional systems, potentially due to the large dimensionality involved in computing the Kalman gain.}, ultimately suffers from divergence during training.
These results validate that LLM-Filter is capable of effectively addressing high-dimensional chaotic systems.

\textbf{Running Time.}
The running time is a crucial factor in practical filtering applications. For learning-based filters and online Bayes filters, we selected RStateNet and EnKF, respectively, as representative methods for comparison. The results are summarized in Figure~\ref{fig.runtime}.
For all systems, the running time of the LLM-Filter is slower than RStateNet, which has fewer parameters. However, LLM-Filter is faster than the online EnKF, making it a highly practical and effective solution for real-time applications.
As the system dimension increases, both EnKF and RStateNet show a significant increase in computation time. 
In contrast, LLM-Filter maintains a stable runtime of around $1.0ms$, which highlights the LLM-Filter’s efficiency in high-dimensional system estimation, offering a clear advantage for such applications.
 Detailed information about the testing equipment configuration, model parameters, and peak memory can be found in Appendix~\ref{apd.experiment_details} and Table~\ref{tab.parameter_and_memory}.


\subsection{Generalization Evaluation for Estimation Task}
\label{subsec.zero-shot_estimation}
In this experiment, we assess the cross-adaptation capability of our proposed LLM-Filter—specifically, its performance when applied to a modified system or even an entirely different system. 

\begin{figure*}[ht]
\begin{center}
\centerline{\includegraphics[width=0.99\textwidth]{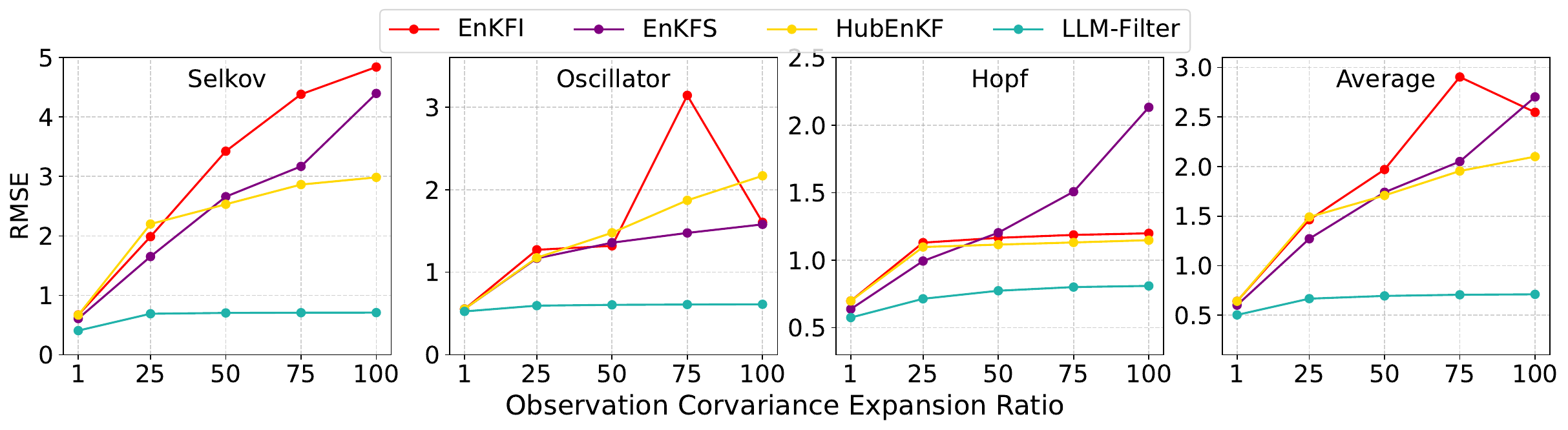}}
\caption{Comparison of RMSE Results across the Selkov, Oscillator, and Hopf systems as the observation covariance expansion ratio increases. 
The rightmost subfigure presents the average results across these systems.
Complete results are presented in Table~\ref{tab.cross_outliers}.}
\label{fig.cross_obs_noise}
\end{center}
\vspace{-20pt}
\end{figure*}

\textbf{Model Mismatch.}
Model mismatch, a common challenge in filtering tasks, occurs when the assumed observation noise variance differs from the actual noise variance \parencite{revach2022kalmannet,duran2024outlier}.
We evaluated LLM-Filter under model mismatch on the Selkov, Oscillator, and Hopf systems, using the observation covariance expansion ratio (OCER) to quantify mismatch severity.
We compared LLM-Filter with the robust filters designed for model mismatch, including EnKFI, EnKFS, and HubEnKF.
The results, shown in Figure~\ref{fig.cross_obs_noise}.
The performance of the robust filters rapidly declined as the OCER increased, while LLM-Filter maintained consistent accuracy with minimal degradation.
These findings underscore the LLM-Filter's exceptional robustness and adaptability with varying observation conditions.
 


\begin{figure*}[ht]
\begin{center}
\centerline{\includegraphics[width=0.99\textwidth]{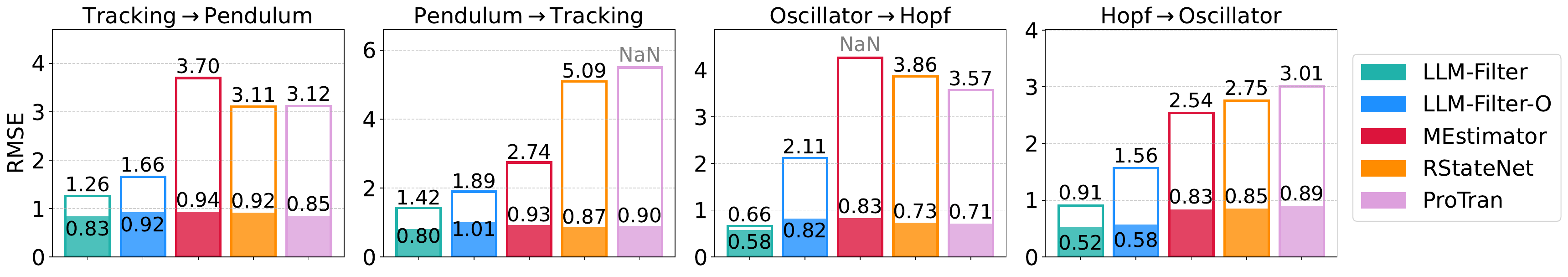}}
\caption{RMSE  for Cross-System Scenarios.
The \emph{hollow bars} represent cross-system scenarios, while the \emph{solid bars} correspond to cases where the model is both trained and tested on the same system.
In cross-system scenarios, \emph{Tracking $\to$ Pendulum} indicates that the model is trained on the Tracking system and evaluated on the Pendulum system.
Comprehensive results are presented in Tab~\ref{tab.zero-shot}.}
\label{fig.zero-shot}
\end{center}
\vspace{-10pt}
\end{figure*}

\textbf{Cross Systems.}
In this experiment, we evaluate the performance on completely different systems, presented in Figure~\ref{fig.zero-shot}.
The cross-system results are shown with \emph{hollow bars}, while \emph{solid bars} represent the performance of baseline models trained directly on the target systems.
We also include the LLM-Filter-O as an ablation comparison.
Guided by the SaP prompt, LLM-Filter achieves exceptional performance, demonstrating the lowest RMSE across all methods in cross-system scenarios.
When compared with specific training baselines, LLM-Filter delivers competitive performance, even achieving the best results in the \emph{Oscillator $\to$ Hopf} scenario.
Without the guidance of the SaP prompt, LLM-Filter-O shows a noticeable decrease in its generalization capability compared to LLM-Filter.
Moreover, the minimal difference between the solid and hollow bars for LLM-Filter indicates that it maintains strong performance even when applied to previously unseen systems.
In contrast, the other learning-based methods experience a significant decline in performance, even losing their ability to provide accurate estimates when confronted with unseen systems.
These results demonstrate the potential of LLM-Filter to generalize effectively across new systems.


\subsection{Exploration of LLMs}
\label{subsec.analysis_llm}
In this subsection, we explore the scaling law and fine-tuning methods of LLMs for LLM-Filter. Additionally, we conduct an ablation study on the LLM component to emphasize its crucial role in achieving the model’s success.

\begin{wrapfigure}{r}{0.45\textwidth}
\centering \includegraphics[width=0.45\textwidth]{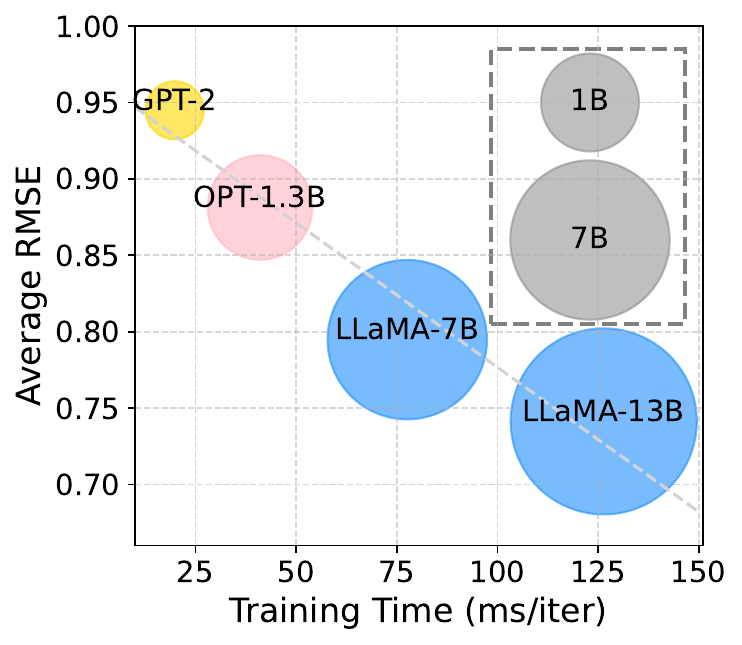}
\caption{Comparison of RMSE and Efficiency Across Various LLMs. Complete results can be found in Table~\ref{tab.various_llms}.}
\label{fig.param_count}
\end{wrapfigure}

\textbf{Scaling Behavior.}
Scaling laws represent empirical observations about how performance and efficiency metrics vary as model size increases \parencite{kaplan2020scaling}. In our study, we explore the scaling behavior of LLM-Filter in state estimation tasks by evaluating models of different parameter sizes as the LLM backbone. 
To avoid inconsistencies caused by varying tokenizers across models, we did not use in-context prompting in this experiment.
We examined the average RMSE and training efficiency across all low-dimensional nonlinear systems and high-dimensional chaotic systems, with results illustrated in Figure~\ref{fig.param_count}.
Our findings illustrate the scaling behavior of LLM-Filter: as model parameters increase, RMSE decreases, boosting estimation accuracy, though at the cost of longer training times. 
Furthermore, our experiments underscore the adaptability of LLM-Filter, as it delivers strong performance across a variety of LLM bases.


\textbf{Full Fine-Tuning and LoRA.}
This experiment investigates two prominent methods commonly used to adapt LLMs more effectively for downstream tasks: full parameter fine-tuning \parencite{dodge2020fine} and low-rank adaptation (LoRA) \parencite{hu2021lora}. For LLM-Filter, full fine-tuning involves updating all model parameters, whereas LoRA only updates a low-rank matrix within the model.
A summary of the results is shown in Table~\ref{tab:lora_finetune}. 
Interestingly, equipping LLM-Filter with LoRA led to a decline in performance. We attribute this to the limitations of low-rank matrix adjustments, which may lack the expressive power needed for the high precision required in state estimation tasks. 
On the other hand, a fully fine-tuned LLM-Filter demonstrated improved performance in more challenging environments, such as Lorenz96, and VL20 systems. This comes at a cost: in simpler dynamic systems, the performance was worse, likely due to overfitting from extensive parameter updates. 


\begin{table}[!t]
\centering
\begin{minipage}[t]{0.45\textwidth}
\centering
\caption{RMSE for LLM-Filter Using LoRA or Full Fine-Tuning.}
\resizebox{\textwidth}{!}{
\begin{tabular}{lccc}
\toprule
\textbf{Method} & \textbf{LLM-Filter} & \textbf{+LoRA} & \textbf{+Full Finetune} \\
\midrule
\textbf{Selkov}    & \textbf{0.4061} & \underline{0.4498} & 0.7103 \\
\textbf{Oscillator} & \textbf{0.5247} & \underline{0.5370} & 0.5513 \\
\textbf{Hopf}      & \textbf{0.5751} & \underline{0.6744} & 0.8444 \\
\textbf{Lorenz96}  & 0.9149 & \underline{0.9147} & \textbf{0.9140} \\
\textbf{VL20}      & \underline{0.7717} & 0.8962 & \textbf{0.7027} \\
\bottomrule
\end{tabular}
}
\label{tab:lora_finetune}
\end{minipage}
\hfill
\begin{minipage}[t]{0.50\textwidth}
\centering
\caption{RMSE for LLM Ablation Study of LLM-Filter.}
\resizebox{\textwidth}{!}{
\begin{tabular}{lcccc}
\toprule
\textbf{Method} & \textbf{LLM-Filter} & \textbf{MLP} & \textbf{RNN} & \textbf{Transformer} \\
\midrule
\textbf{Selkov}      & \textbf{0.4061} & 0.9910 & 0.9862 & \underline{0.8863} \\
\textbf{Oscillator}  & \textbf{0.5247} & \underline{0.8176} & 0.9481 & 0.8566 \\
\textbf{Hopf}        & \textbf{0.5751} & 0.9057 & 1.0114 & \underline{0.8610} \\
\textbf{Lorenz96}    & \textbf{0.9149} & \underline{0.9661} & 1.0105 & 0.9721 \\
\textbf{VL20}        & \textbf{0.7717} & 0.9977 & \underline{0.8043} & 0.9598 \\
\bottomrule
\end{tabular}
}
\label{tab:llm_ablation}
\end{minipage}
\vspace{-10pt}
\end{table}

\textbf{LLM Ablation Study.}
In this experiment, we address whether the use of LLMs is truly essential for state estimation tasks, a concern raised in recent studies on LLMs for time series analysis \parencite{tan2024language}. To investigate this, we replace the LLM in LLM-Filter with traditional neural network architectures, including MLP, RNN, and Transformer, while keeping all other components unchanged. The results are summarized in Table~\ref{tab:llm_ablation}.
Our findings clearly demonstrate that LLM-Filter consistently outperforms these traditional network architectures across all tested systems. Specifically, the RMSE of LLM-Filter shows an average improvement of $31.66\%$, $32.43\%$, and $30.73\%$ over MLP, RNN, and Transformer, respectively. 
This significant improvement underscores the superior performance and essential role of LLMs in state estimation tasks, validating their necessity for achieving state-of-the-art results.


\section{Conclusion and Future Work} \label{sec.conclusion}
Inspired by the success of large control models, we propose a general large filtering model, LLM-Filter, designed to solve specific estimation tasks, including generalization tasks, by transferring knowledge from LLMs. The process begins by embedding noisy observations with text prototypes and then enriching the input with the SaP method, which incorporates basic task instructions and characteristics. These inputs are then processed by a frozen LLM. The generated outputs are projected onto the final estimates.
Guided by prompts and leveraging pretraining knowledge, LLM-Filter outperforms specialized learning-based methods and demonstrates strong generalization across various systems. 

The generalization ability of LLM-Filter is currently restricted to systems with the same dimensionality, and its evaluation across diverse settings is still limited. 
Future work will focus on training large-scale state estimation models with multi-source data, aiming to enable generalization across systems with varying dimensions. 
We hope that our contributions will lay the groundwork for addressing state estimation challenges in a wide range of real-world applications.

\section*{Acknowledgements}
This study is supported by Tsinghua-Efort Joint Research Center for EAI Computation and Perception.



{
\small
\printbibliography
}


\newpage
\appendix

\section{Bayes Filters} 
\label{apd.filter_methods}

In this section, we introduce the classical Bayes filters, including Kalman filter (KF), Extended Kalman filter (EKF), Ensemble Kalman filter (EnKF), and Particle filter (PF). For detailed descriptions of advanced variants, readers are encouraged to explore the relevant literature cited in Section \ref{sec.experiments}.

\subsection{Kalman Filter}
Suppose the SSM is linear and Gaussian:
\begin{equation} \nonumber
\begin{aligned}
\boldsymbol{x}_{t+1}&=\boldsymbol{F}_t\boldsymbol{x}_{t}+\boldsymbol{\xi}_{t}, \\
\boldsymbol{y}_{t}&=\boldsymbol{H}_t\boldsymbol{x}_{t}+\boldsymbol{\zeta}_{t},
\end{aligned}
\end{equation}
where $\boldsymbol{\xi}_t\sim \mathcal{N}(\boldsymbol{\xi}_t|0, \boldsymbol{Q_t})$ and $\boldsymbol{\zeta}_t\sim \mathcal{N}(\boldsymbol{\zeta}_t|0, \boldsymbol{R_t})$ represent zero-mean Gaussian noise terms with covariance matrices $\boldsymbol{Q_t}\in \mathbb{R}^{M\times M}$ and $\boldsymbol{R_t}\in \mathbb{R}^{N\times N}$, respectively. The initial state is a Gaussian distribution, given by $\boldsymbol{x}_0\sim \mathcal{N}(\boldsymbol{x}_0|\boldsymbol{\mu}_0, \boldsymbol{\Sigma_0})$
with known mean $\boldsymbol{\mu}_0\in \mathbb{R}^{M}$ and covariance $\boldsymbol{\Sigma_0}\in \mathbb{R}^{M\times M}$. 
According to the Kalman filter \parencite{kalman1960new},
the prior predictive distribution and posterior distribution remain Gaussian:
\begin{equation}
\begin{aligned}
\nonumber
p(\boldsymbol{x}_t|\boldsymbol{y}_{1:t-1})& =\mathcal{N}(\boldsymbol{x}_t\mid\boldsymbol{\mu}_{t|t-1}, \boldsymbol{\Sigma}_{t|t-1}), \\
p(\boldsymbol{x}_t|\boldsymbol{y}_{1:t})& =\mathcal{N}(\boldsymbol{x}_t\mid\boldsymbol{\mu}_t, \boldsymbol{\Sigma}_t). 
\end{aligned}
\end{equation}
The mean and covariance of the prior predictive distribution are denoted by:
\begin{equation}\begin{aligned}
\nonumber
\boldsymbol{\mu}_{t|t-1}&=\boldsymbol{F}_t\boldsymbol{\mu}_{t-1},\\
\boldsymbol{\Sigma}_{t|t-1}&=\boldsymbol{F}_t\boldsymbol{\Sigma}_{t-1}\boldsymbol{F}_t^\top+\boldsymbol{Q}_t,\end{aligned}\end{equation}
The posterior mean and covariance are given by:
\begin{equation}
\begin{aligned}
\nonumber
\boldsymbol{\Sigma}_{t}^{-1}& =\boldsymbol{\Sigma}_{t|t-1}^{-1}+\boldsymbol{H}_t^\top\boldsymbol{R}_t^{-1}\boldsymbol{H}_t, \\
\mathbf{K}_{t}& =\boldsymbol{\Sigma}_t\boldsymbol{H}_t^\top\boldsymbol{R}_t^{-1}, \\
\boldsymbol{\mu}_{t}& =\boldsymbol{\mu}_{t|t-1}+\mathbf{K}_t\left(\boldsymbol{y}_t-\boldsymbol{H}_t\boldsymbol{\mu}_{t|t-1}\right). 
\end{aligned}
\end{equation}
where $\boldsymbol{K}_t$ is the Kalman gain matrix.

\subsection{Extended Kalman Filter}
When faced with nonlinear systems, EKF introduces the first-order Taylor approximations of the nonlinear transition model $f(\cdot)$ and observation model $h(\cdot)$.
We denote the Jacobians of these functions as $\boldsymbol{F_x}(\cdot)$ and $\boldsymbol{H_x}(\cdot)$. Then the predict step of EKF is:
\begin{equation}\begin{aligned}
\nonumber
\boldsymbol{\mu}_{t|t-1}&=f(\boldsymbol{\mu}_{t-1}),\\
\boldsymbol{\Sigma}_{t|t-1}&=\boldsymbol{F}_x(\boldsymbol{\mu}_{t-1})\boldsymbol{\Sigma}_{t-1}\boldsymbol{F}_x(\boldsymbol{\mu}_{t-1})^\top+\boldsymbol{Q}_t.
\end{aligned}
\end{equation}
The update step is 
\begin{equation}
\begin{aligned}
\nonumber
\boldsymbol{\Sigma}_{t}^{-1}& =\boldsymbol{\Sigma}_{t|t-1}^{-1}+\boldsymbol{H}_x(\boldsymbol{\mu}_{t|t-1})^\top\boldsymbol{R}_t^{-1}\boldsymbol{H}_x(\boldsymbol{\mu}_{t|t-1}), \\
\mathbf{K}_{t}& =\boldsymbol{\Sigma}_t\boldsymbol{H}_x(\boldsymbol{\mu}_{t|t-1})^\top\boldsymbol{R}_t^{-1}, \\
\boldsymbol{\mu}_{t}& =\boldsymbol{\mu}_{t|t-1}+\mathbf{K}_t\left(\boldsymbol{y}_t-\boldsymbol{H}_x(\boldsymbol{\mu}_{t|t-1})\boldsymbol{\mu}_{t|t-1}\right). 
\end{aligned}
\end{equation}

\subsection{Ensemble Kalman Filter}
The EnKF was developed as an alternative to the EKF for high-dimensional and chaotic systems, such as weather forecasting \parencite{carrassi2018data}.
Instead of explicitly computing the covariance matrix, EnKF represents the belief state with an ensemble of $N_e$ particles ${\hat{\boldsymbol{x}}_t^i}\in \mathbb{R}^M$ , where $i=1,\cdots,N_e$.
The predict step of EnKF obtains ${\hat{\boldsymbol{x}}_{t|t-1}^i}$ using the transition model from  \eqref{eq.prediction}.
In the update step, predicted observations  ${\hat{\boldsymbol{y}}_t^i}$ are sampled for each particle as: 
\begin{equation}\begin{aligned}
\nonumber
\hat{\boldsymbol{y}}_t^i\sim \mathcal{N}(\hat{\boldsymbol{y}}_t^i|h(\hat{\boldsymbol{x}}_{t|t-1}^i), \boldsymbol{R}_t) \quad i=1,2,\cdots,N_e.
\end{aligned}
\end{equation}
Next, the particles are updated as follows: 
\begin{equation}\begin{aligned}
\nonumber
\boldsymbol{K}_t &= \mathrm{Cov}( \hat{\boldsymbol{x}}_{t|t-1}^i,\hat{\boldsymbol{y}}_t^i) \mathbb{V}( \hat{\boldsymbol{y}}_t^i)^{-1}, \\
\hat{\boldsymbol{x}}_{t}^i &= \hat{\boldsymbol{x}}_{t|t-1}^i + \boldsymbol{K}_t(\boldsymbol{y}_t-\hat{\boldsymbol{y}}_t^i),\ i=1,2,\cdots,N_e,
\end{aligned}
\end{equation}
where $\mathrm{Cov}(\cdot,\cdot)$ and  $\mathbb{V}(\cdot)$ denote the covariance and variance, respectively. Finally, the state estimate at time $t$ is calculated as the average of all particles:
\begin{equation}\begin{aligned}
\nonumber
\hat{\boldsymbol{x}}_{t}= \sum_{i=1}^{N_e}\boldsymbol{x}^i_{t}.
\end{aligned}
\end{equation}

\subsection{Particle Filter}
The PF is a filtering method based on sequential importance sampling \parencite{liu1998sequential}, characterized by high precision but relatively low computational efficiency. The standard implementation of PF follows the bootstrap procedure outlined below: 

(1) \textbf{Sampling:} Generate samples for all $N_p$ particles;
\begin{equation}\begin{aligned}
\nonumber
\boldsymbol{x}^i_{t}\sim p(\boldsymbol{x}_{t}|\boldsymbol{x}^i_{t-1}), \quad i=1,2,\cdots,N_p.
\end{aligned}
\end{equation}
(2) \textbf{Weighting:} compute weights of all particles and normalize;
\begin{equation}\begin{aligned}
\nonumber
w^i_{t}\propto   p(\boldsymbol{y}_{t}|\boldsymbol{x}^i_{t}), \quad i=1,2,\cdots,N_p.
\end{aligned}
\end{equation}
(3) \textbf{Resampling}: Resample particles based on their normalized weights. 

Finally, the state estimate at time $t$ is calculated as the weighted average of all particles: 
\begin{equation}\begin{aligned}
\nonumber
\hat{\boldsymbol{x}}_{t}= \sum_{i=1}^{N_p}w^i_{t}\boldsymbol{x}^i_{t}.
\end{aligned}
\end{equation}

\begin{table}[htbp]
  \caption{System Specifications for Experiments.  
  The dimensions of the state and observation spaces are denoted by \( M \) and \( N \), respectively. \emph{Dataset Size} refers to the collected data from the system, specified as (trajectory length, number of trajectories, and state dimension). 
  The \emph{Forward Euler} method refers to a first-order explicit discretization scheme, while \emph{RK4} represents the fourth-order Runge-Kutta discretization scheme \parencite{butcher1964implicit}.}
  \label{tab.system_specifications}
  \vskip 0.05in
  \footnotesize
  \begin{threeparttable}
  \renewcommand\arraystretch{1.4}
  \setlength{\tabcolsep}{6pt}
  \resizebox{\textwidth}{!}{\begin{tabular}{c|c|c|c|c|c|c|c|c|c}
    \toprule
    \textbf{System} & \textbf{Linear} & \textbf{M} & \textbf{N} & \textbf{Dataset Size} & \(\boldsymbol{Q}\) & \(\boldsymbol{R}\) & \( \Delta t \) & \textbf{Discretization} & \textbf{Application Domain} \\
    \toprule
    Tracking & T & 4 & 2 & (200, 100, 6) & $0.1\boldsymbol{I}$ & $10\boldsymbol{I}$ & 0.1 & Forward Euler & Target Tracking \\
    \midrule
    Selkov  & F & 2 & 2 & (200, 100, 4) & $\boldsymbol{I}$ & $\boldsymbol{I}$ & 0.01 & RK4 & Glycolysis Process  \\
    \midrule
    Oscillator & F & 2 & 2 & (200, 100, 4) & $\boldsymbol{I}$ & $\boldsymbol{I}$ & 0.01 & RK4 & Oscillatory Motion  \\
    \midrule
    Hopf  & F & 2 & 2 & (200, 100, 4) & $\boldsymbol{I}$ & $\boldsymbol{I}$ & 0.01 & RK4 & Chemical Reactions \\
    \midrule
    Pendulum & F & 4 & 2 & (200, 100, 6) & $\boldsymbol{I}$ & $\boldsymbol{I}$ & 0.01 & RK4  & Physical Dynamics \\
    \midrule
    Lorenz 96 & F & 72 & 72 & (200, 100, 144) & $\boldsymbol{I}$ & $100\boldsymbol{I}$ & 0.01 & RK4 & Atmospheric Dynamics \\
    \midrule
    VL20 & F & 72 & 72 & (200, 100, 144) & $\boldsymbol{I}$ & $\boldsymbol{I}$ & 0.01 & RK4 & Atmospheric Dynamics \\
    \bottomrule
  \end{tabular}}
  \end{threeparttable}
\end{table}

\begin{figure}[ht]
    \centering
     \resizebox{0.95\textwidth}{!}
     {
    \begin{subfigure}{0.2\textwidth}
        \centering
        \includegraphics[width=\linewidth]{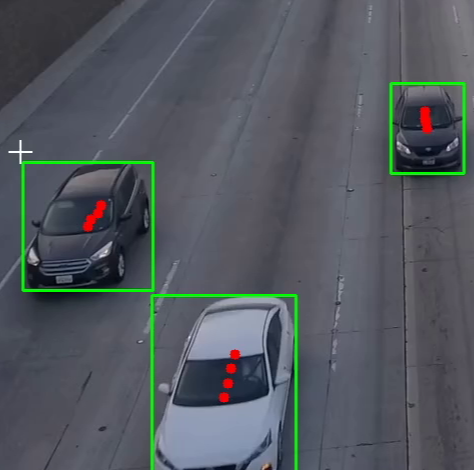} 
        \caption{}
        \label{fig:tracking}
    \end{subfigure}
    \hspace{0.03\textwidth}
    \begin{subfigure}{0.2\textwidth}
        \centering
        \includegraphics[width=\linewidth]{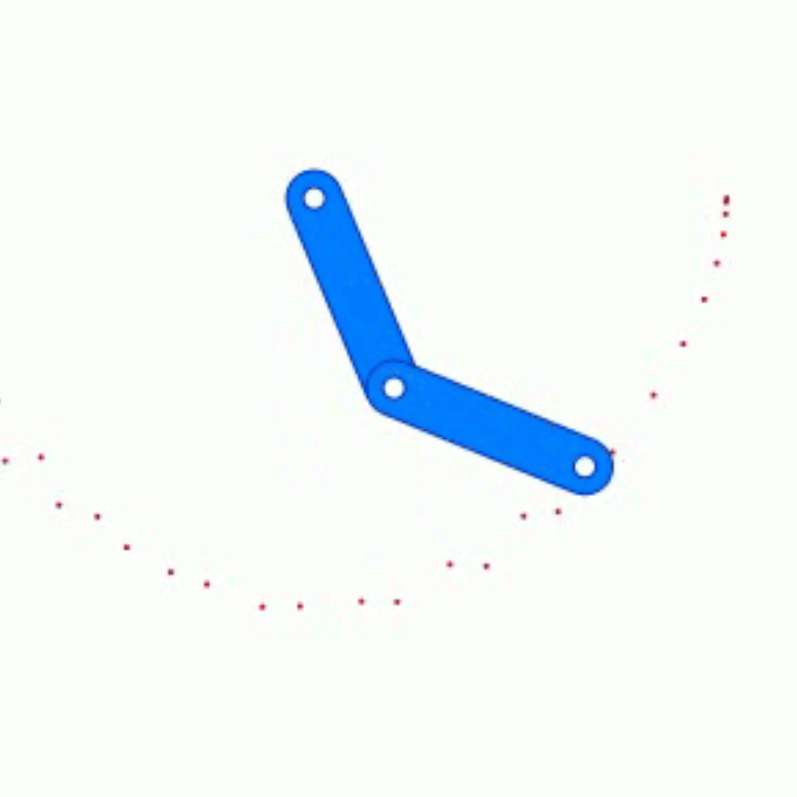}
        \caption{}
        \label{fig.pendulum}
    \end{subfigure}
    \hspace{0.03\textwidth}
    \begin{subfigure}{0.2\textwidth}
        \centering
        \includegraphics[width=\linewidth]{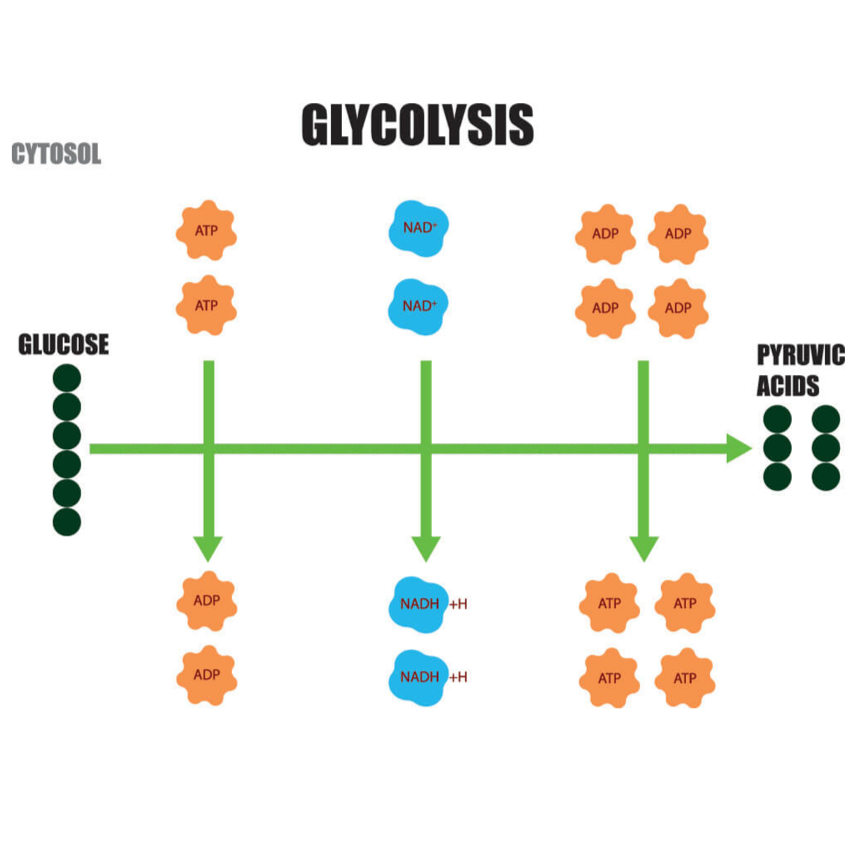}
        \caption{}
        \label{fig.Selkov}
    \end{subfigure}

    \vspace{0.3cm} 

    \begin{subfigure}{0.2\textwidth}
        \centering
        \includegraphics[width=\linewidth]{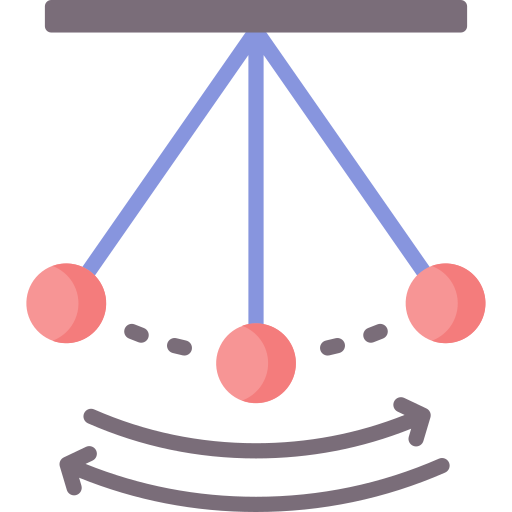}
        \caption{}
        \label{fig.Oscillator}
    \end{subfigure}
    \hspace{0.03\textwidth}
    \begin{subfigure}{0.2\textwidth}
        \centering
    \includegraphics[width=\linewidth]{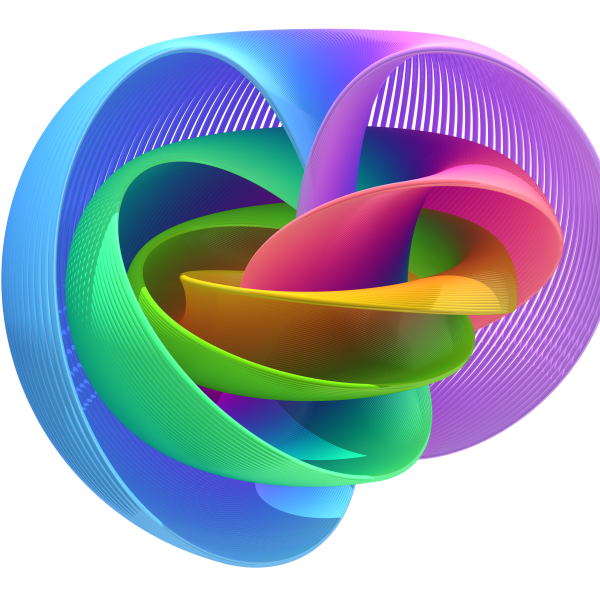}
        \caption{}
        \label{fig.Hopf}
    \end{subfigure}
    \hspace{0.03\textwidth}
    \begin{subfigure}{0.2\textwidth}
        \centering
        \includegraphics[width=\linewidth]{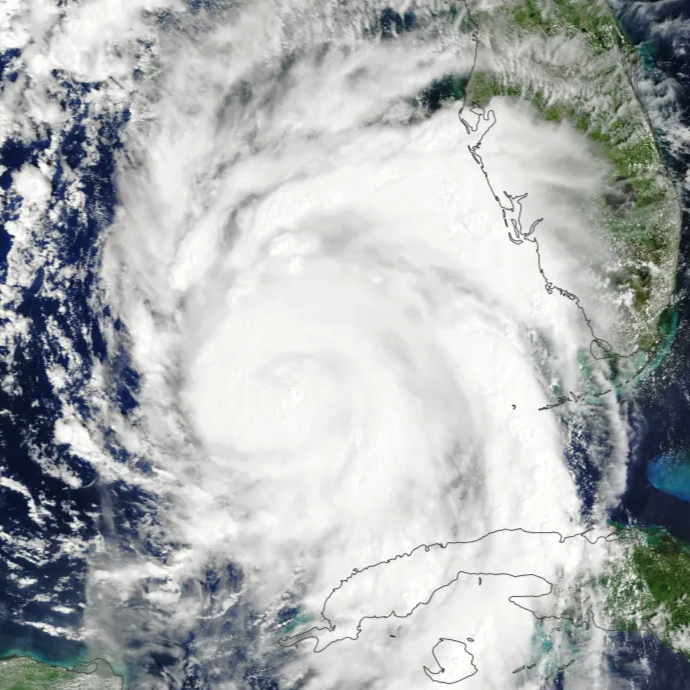}
        \caption{}
        \label{fig.atmospher}
    \end{subfigure}
    }
\caption{Illustration of Experiment Systems. (a) Tracking; (b) Double Pendulum; (c) Selkov; (d) Oscillator; (e) Hopf; (f) Lorenz96 \& VL20.}
    \label{fig.systems}
\end{figure}

\section{System Descriptions} \label{apd.system_description}
The systems used in this experiment are based on real-world models that are continuous in nature and governed by ordinary differential equations (ODEs). To map these continuous systems into the corresponding Markovian state-space model (SMM), we discretize them using methods such as the forward Euler method and the fourth-order Runge-Kutta discretization \parencite{butcher1964implicit}. For the uncertain noise terms in \eqref{eq.ssm}, denoted as $\boldsymbol{\xi}_t$ and $\boldsymbol{\zeta}_t$, we assume they follow a standard Gaussian distribution, ensuring consistency across the models.
\begin{equation}
\begin{aligned}
\nonumber
\boldsymbol{\xi}_t \sim \mathcal{N}(\boldsymbol{0},  \boldsymbol{Q}), \ 
\boldsymbol{\zeta}_t \sim \mathcal{N}(\boldsymbol{0}, \boldsymbol{R}),
\end{aligned}
\end{equation}
where $\boldsymbol{Q}\in \mathbb{R}^{M\times M}$ and $\boldsymbol{R}\in \mathbb{R}^{N\times N}$ are the covariances.
A summary of the essential characteristics of all the systems is provided in Table~\ref{tab.system_specifications} and Figure~\ref{fig.systems}. Following this, the specific background and mathematical modeling for each system are discussed in detail.

\textbf{Tracking.} The Tracking system, also known as the Wiener velocity model, is a canonical linear Gaussian system \parencite{sarkka2023bayesian} commonly used for target Tracking \parencite{weng2020ab3dmot}. The state represents the position and velocity of a moving object in two dimensions, expressed as $x = \left[p_x \ p_y \ v_x \ v_y \right]^{\top}$, where $p_x$ and $p_y$ denote the object's positions in the longitudinal and lateral directions, respectively, and $v_x$ and $v_y$ are the corresponding velocities. 
The observations are direct, noisy observations of the position components.
The overall system can be described as
\begin{equation}
\nonumber
\begin{aligned}
\dot{\boldsymbol{x}}_{} &= \begin{bmatrix}
0 & 0 & 1 & 0 \\
0 & 0 & 0 & 1 \\
0 & 0 & 0 & 0 \\
0 & 0 & 0 & 0
\end{bmatrix} \boldsymbol{x},
\\
\boldsymbol{y}_t &= \begin{bmatrix}
1 & 0 & 0 & 0 \\
0 & 1 & 0 & 0
\end{bmatrix} \boldsymbol{x}_t+\boldsymbol{\zeta}_t .
\end{aligned}
\end{equation}

\textbf{Selkov.} The Selkov system models the dynamics of the glycolysis process through a set of two coupled differential equations \parencite{sel1968self}. The state vector \( \boldsymbol{x} = \left[\theta_1 \ \theta_2 \right]^{\top} \) represents the concentrations of the metabolites involved in the glycolytic cycle, where \( \theta_1 \) and \( \theta_2 \) correspond to the concentrations of the key metabolites. The constants \( a=0.08 \) and \( b=0.6 \) are parameters related to the reaction rates within the system.
The dynamics of the system and the observation process are governed by the following equations:
\begin{equation}
\nonumber
\begin{aligned}
\dot{\boldsymbol{x}} &= \begin{bmatrix}
\dot{\theta_1} \\
\dot{\theta_2}
\end{bmatrix}
=
\begin{bmatrix}
-\theta_1 + a \theta_2 + \theta_1^2 \theta_2 \\
b - a \theta_2 - \theta_1^2 \theta_2
\end{bmatrix}, \quad
\boldsymbol{y}_t = \boldsymbol{x}_t + \boldsymbol{\zeta}_t.
\end{aligned}
\end{equation}


\textbf{Damped Oscillator.} The damped Oscillator refers to a physical system in which an object experiences Oscillatory motion that gradually decreases over time due to the presence of a damping force \parencite{brunton2016discovering}. In this system, the state vector \( \boldsymbol{x} = \left[\theta_1 \ \theta_2\right]^{\top} \) represents the positions of the Oscillator along two axes, \( \theta_1 \) and \( \theta_2 \).
 The overall system can be described as:
\begin{equation}
\nonumber
\begin{aligned}
\dot{\boldsymbol{x}} &= \begin{bmatrix}
\dot{\theta_1} \\
\dot{\theta_2}
\end{bmatrix}
=
\begin{bmatrix}
-a \theta_1^3 + b \theta_2^3 \\
-b \theta_1^3 - a \theta_2^3
\end{bmatrix},\quad
\boldsymbol{y}_t = \boldsymbol{x}_t + \boldsymbol{\zeta}_t.
\end{aligned}
\end{equation}
Here, \( a=0.1 \) and \( b=2 \) are constants that define the strength of the nonlinearity, and the cubic terms represent the restoring forces in the system. 


\textbf{Hopf Bifurcation.} The Hopf bifurcation describes a dynamical system that undergoes spontaneous oscillations as a system parameter \( \mu \) crosses a critical threshold, transitioning from a stable equilibrium to an Oscillatory regime \parencite{hassard1981theory}. This phenomenon is commonly observed in various real-world systems, such as chemical reactions, electrical circuits, and fluid dynamics.
In this system, the state vector is represented as \( \boldsymbol{x} = \begin{bmatrix} \theta_1 \ \theta_2 \end{bmatrix}^{\top} \). The parameter \( \mu=0.5 \) controls the stability of the system, while \( \omega=1 \) and \( A=1 \) are constants that define the frequency of oscillations and the strength of the nonlinear interactions, respectively.
The overall system can be described as
\begin{equation}
\nonumber
\begin{aligned}
\dot{\boldsymbol{x}} &= \begin{bmatrix}
\dot{\theta_1} \\
\dot{\theta_2}
\end{bmatrix}
=
\begin{bmatrix}
\mu \theta_1 + \omega \theta_2 - A \theta_1 (\theta_1^2 + \theta_2^2) \\
-\omega \theta_1 + \mu \theta_2 - A \theta_2 (\theta_1^2 + \theta_2^2)
\end{bmatrix}, \quad
\boldsymbol{y}_t = \boldsymbol{x}_t + \boldsymbol{\zeta}_t.
\end{aligned}
\end{equation}

\textbf{Double Pendulum.} The Double Pendulum, often referred to as the chaotic pendulum, is a nonlinear dynamical system consisting of a pendulum with another pendulum attached to its end \parencite{levien1993double}. The state vector \( \boldsymbol{x} = \left[\theta_1 \ \omega_1 \ \theta_2 \ \omega_2 \right]^{\top} \) encapsulates the system's configuration, where \( \theta_1 \) and \( \theta_2 \) represent the angular positions, and \( \omega_1 \) and \( \omega_2 \) denote the angular velocities of the two pendulums. 
The system's observations are the noisy observations of the angles \( \theta_1 \) and \( \theta_2 \):
\begin{equation}
\nonumber
\begin{aligned}
\dot{\boldsymbol{x}} &= \begin{bmatrix}
\dot{\theta_1} \\
\dot{\omega_1} \\
\dot{\theta_2} \\
\dot{\omega_2}
\end{bmatrix}
=
\begin{bmatrix}
\omega_1 \\
\frac{M_2 L_1 \omega_1^2 \sin(D) \cos(D) + M_2 G \sin(\theta_2) \cos(D) + M_2 L_2 \omega_2^2 \sin(D) - (M_1 + M_2) G \sin(\theta_1)}{(M_1 + M_2) L_1 - M_2 L_1 \cos^2(D)} \\
\omega_2 \\
\frac{-M_2 L_2 \omega_2^2 \sin(D) \cos(D) + (M_1 + M_2) G \sin(\theta_1) \cos(D) - (M_1 + M_2) L_1 \omega_1^2 \sin(D) - (M_1 + M_2) G \sin(\theta_2)}{\frac{L_2}{L_1} \left( (M_1 + M_2) L_1 - M_2 L_1 \cos^2(D) \right)}
\end{bmatrix},
\\
\boldsymbol{y}_t &= \begin{bmatrix}
1 & 0 & 0 & 0 \\
0 & 0 & 1 & 0
\end{bmatrix} \boldsymbol{x}_t + \boldsymbol{\zeta}_t, 
\quad D = \theta_2 - \theta_1.
\end{aligned}
\end{equation}
Here, \( M_1=1 \) and \( M_2=1 \) are the masses of the two pendulums, \( L_1=1 \) and \( L_2=1 \) represent their respective lengths, and \( G=9.8 \) is the gravitational constant that governs the motion of the system.


\textbf{Lorenz96.}
The Lorenz96 system, conceived by Edward Lorenz \parencite{lorenz1996predictability}, offers a simplified yet powerful model of atmospheric dynamics, often serving as a cornerstone in chaos theory and meteorology. This system is described by a set of ordinary differential equations that capture the intricate interplay between atmospheric modes in a periodic domain:

\begin{equation} \nonumber\begin{aligned} \dot{\boldsymbol{x}}^{(j)} & = (\boldsymbol{x}^{(j+1)} - \boldsymbol{x}^{(j-2)})\boldsymbol{x}^{(j-1)} - \boldsymbol{x}^{(j)} + F, \quad \boldsymbol{y}_t = \boldsymbol{x}_t + \boldsymbol{\zeta}_t, \quad j = 1, 2, \dots, M. \end{aligned} \end{equation}

In this model, $\boldsymbol{x}^{(j)}$ represents the $j$th state of the Lorenz96 system, while $F=8$ is the external forcing term that drives the system. The system operates under periodic boundary conditions, ensuring continuity across the domain: $\boldsymbol{x}^{(M+1)} = \boldsymbol{x}^{(1)}$, $\boldsymbol{x}^{(0)} = \boldsymbol{x}^{(M)}$, and $\boldsymbol{x}^{(-1)} = \boldsymbol{x}^{(M-1)}$, with $M=72$ representing the total number of dimensions in the state space.


\textbf{Vissio-Lucarini 20.}
The Vissio-Lucarini 20 model (VL20), introduced by Vissio and Lucarini \parencite{vissio2020mechanics}, presents a coupled system that provides a minimalist yet rich representation of Earth's atmospheric dynamics. By extending the Lorenz framework to include \emph{temperature} variables, the VL20 model enables the development of more complex and nuanced behavioral patterns:
\begin{equation} \nonumber \begin{aligned} \dot{\boldsymbol{\phi}}^{(j)} & = (\boldsymbol{\phi}^{(j+1)} - \boldsymbol{\phi}^{(j-2)})\boldsymbol{\phi}^{(j-1)} - \gamma \boldsymbol{\phi}^{(j)} - \varepsilon \boldsymbol{\theta}^{(j)} + F, \\ \dot{\boldsymbol{\theta}}^{(j)} & = \boldsymbol{\phi}^{(j+1)} \boldsymbol{\theta}^{(j+2)} - \boldsymbol{\phi}^{(j-1)} \boldsymbol{\theta}^{(j-2)} + \varepsilon \boldsymbol{\phi}^{(j)} - \gamma \boldsymbol{\theta}^{(j)} + G, \\ \boldsymbol{y}_t & = \boldsymbol{x}_t + \boldsymbol{\zeta}_t, \quad j = 1, 2, \dots, M'. \end{aligned} \end{equation}
In this model, $M'$ represents the number of discrete points in the system, and the state vector is given by $\boldsymbol{x} = \begin{bmatrix} \boldsymbol{\phi} \ \boldsymbol{\theta} \end{bmatrix}^{\top}$, with $M=2M'$ variates. The constants $F=10, G=0, \gamma=1, \varepsilon=1$ are defined in \parencite{vissio2020mechanics}. 
Furthermore, the system is subject to periodic boundary conditions, akin to those in the Lorenz96 model.

\section{Principle Analysis}
\label{apd.Principle Analysis}

\begin{figure*}[ht]
\begin{center}
\centerline{\includegraphics[width=0.8\textwidth]{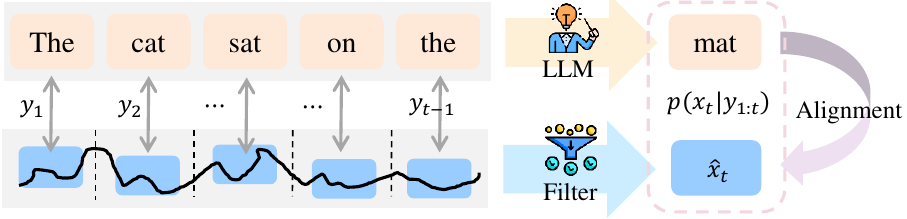}}
\caption{Principle Comparison of LLMs and Filtering. The LLM predicts the next token (e.g., \emph{mat}) based on preceding tokens, while a filter estimates the current state $\hat{x}_t$ using past observations.}
\label{fig.principle}
\end{center}
\vspace{-10pt}
\end{figure*}

In this section, we analyze the underlying connection between large language models (LLMs) and filtering. As illustrated in Figure~\ref{fig.principle}, we present a simplified comparison of the two processes. LLMs predict the probability distribution of the next token based on all preceding tokens, while filtering estimates the distribution of the current system state given all past observations. 
This fundamental similarity, where both approaches make predictions based on sequential historical information, enables a natural alignment between them. With appropriate alignment, LLM-Filter can effectively leverage the reasoning capabilities and rich pre-trained knowledge of LLMs for state estimation tasks.

Regarding a more detailed theoretical explanation, it is challenging to provide one at this stage. This limitation stems from the fact that the theoretical foundations of LLMs themselves remain an open research problem \parencite{achiam2023gpt,openai2023gpt}.Consequently, many applications of LLMs across diverse domains—such as time series forecasting \parencite{liu2024autotimes, zhou2023one, jin2023time} and control systems \parencite{brohan2022rt, brohan2023rt, kim2024openvla}, also lack rigorous mathematical justification. Similarly, our work focuses on empirical evidence and the practical alignment between language modeling and filtering, while acknowledging the need for deeper theoretical understanding in future research.

\section{Experiment Details}
\label{apd.experiment_details}
Our model is built with PyTorch \parencite{paszke2019pytorch} and trained extensively 
on the same server with the following hardware configuration: 4 × Intel® Xeon® Gold 6226R @ 2.90GHz CPUs and 1 × NVIDIA H800-80G GPU.
The complete code will be released after thorough organization and conflict-of-interest review.
\subsection{Benchmark Settings}
\label{apd.benchmar_settings}
The online Bayes baseline methods are implemented by JAX \parencite{jax2018github} following \parencite{duran2024outlier}. 
The number of particles for EnKF, EnKFI, EnKFS, and HubEnKF is set to 1000 to balance efficiency and accuracy \parencite{raanes2024dapper}.
For a fair comparison, the hyperparameters of all online Bayes methods are selected on the first trial using the Bayesian Optimization (BO) package \parencite{gardner2014bayesian} with a parameter range of $(1,20) $. 
The KalmanNet \parencite{revach2022kalmannet} implementation is based on their official open-source repository. 
The MEstimator \parencite{ji2022concurrent}, RStateNet \parencite{dahal2024robuststatenet}, and ProTran \parencite{tang2021probabilistic} are implemented according to the methodologies outlined in their respective papers.
The parameter counts and peak memory for different models in Table~\ref{tab.parameter_and_memory}. Since we froze the core layers of Llama-7B and only trained the ObsEmbedding and StateProjection layers, we distinguish between the total parameter count and the trainable parameter count.

Our experiments are conducted on the systems described in Appendix \ref{apd.system_description}. We gather 100 trajectories, each consisting of 200 time steps, from each system, as shown in Table~\ref{tab.system_specifications}. To prevent data leakage, the datasets are split chronologically into training, validation, and test sets in a $7:1:2$ ratio.
\begin{table}[ht]
\centering 
\caption{Comparison of Model Parameters and Peak Memory}
\label{tab.parameter_and_memory}
\resizebox{0.95\textwidth}{!}{
\begin{tabular}{cccccc}
\toprule
\multirow{2}{*}{\textbf{Model}} & \textbf{LLM-Filter } & \multirow{2}{*}{\textbf{MEstimator}} & \multirow{2}{*}{\textbf{RStateNet}} & \multirow{2}{*}{\textbf{ProTran}} & \multirow{2}{*}{\textbf{KalmanNet}} \\
&\textbf{(Total/Trainable)}\\
\midrule
\textbf{Parameter} & 6.61B / 4.22M & 2.12M & 7.38M & 15.77M & 5.20K \\
\midrule
\textbf{Peak memory} & 27.43GiB & 622.51MiB & 742.57MiB & 854.54MiB & 582.93MiB \\
\bottomrule
\end{tabular}}
\end{table}
\subsection{Implementation} 
\label{apd.implementation}
In LLM-Filter, only the $\operatorname{ObsEmbedding}$and $\operatorname{StateProjection}$
layers are updated via the gradient. These components are implemented using a simple MLP, which has been shown to effectively embed and project time series for LLM  \parencite{liu2024autotimes}.
The default hidden dimension and number of MLP layers, batch size, window length $T$, and segment length $L$ are set to 512, 2, 16 ,40, and 20 respectively. We use the AdamW optimizer \parencite{loshchilov2017fixing} with the initial learning rate of $10^{-4}$ and the weight decay of $10^{-5}$, where the model is trained for 10 epochs.
Unless otherwise specified, we use LLaMA-7B \parencite{touvron2023llama} as the default base LLM.

\subsection{Evaluation Metrics}
\label{apd.metrics}
For evaluation, we use the root mean square error (RMSE) as the primary metric to assess the accuracy of state estimation \parencite{sarkka2023bayesian}. To evaluate efficiency, we measure the inference running time per step in milliseconds (ms/step). The calculations for these metrics are as follows:
\begin{equation}
\nonumber
{\text{RMSE}} =
\sqrt{\frac{\sum_{t=1}^{T_{traj}}\Vert x_t - \hat{x}_t \Vert^2}{T_{traj}}}, \quad
\text{Runtime}=\frac{\text{Cost Time}}{T_{traj}\cdot N_{test}},
\end{equation}
where $T_{traj}$ is the trajectory length and $N_{test}$ denotes the number of trajectories in the test set.


\section{Supplementary Results}
\label{apd.supplementary_results}


\subsection{Running Time}
We evaluated the estimation efficiency (ms/step) of various methods on the same device, as described in Appendix \ref{apd.experiment_details}. The results are presented in Table~\ref{tab.runtime}. 

\begin{table*}[ht!]
\centering
\caption{Comparison of Estimation Runtime (ms/step) Across Methods.}
\resizebox{0.5\textwidth}{!}{
\begin{tabular}{lccc}
\toprule
\textbf{Method} & \textbf{LLM-Filter} & \textbf{EnKF} & \textbf{RStateNet} \\ 
\midrule
\textbf{Selkov}    & 0.8766 & 2.5095 & 0.1546 \\
\textbf{Oscillator} & 0.8299 & 2.6815 & 0.1553 \\
\textbf{Hopf}       & 0.8245 & 2.5747 & 0.1267 \\
\textbf{Pendulum}   & 1.2447 & 2.2246 & 0.0967 \\
\textbf{Lorenz96}   & 0.9577 & 6.7377 & 0.4513 \\
\textbf{VL20}       & 0.9503 & 2.9397 & 0.4521 \\
\bottomrule
\end{tabular}}
\label{tab.runtime}
\end{table*}


\subsection{Generalization Estimation Experiments}
\label{apd.zero-shot}
To assess the generalization capabilities of LLM-Filter, we conducted extensive evaluation. First, we evaluated its performance on systems subjected to varying strengths of outliers, as detailed in Table~\ref{tab.cross_outliers}. Notably, while the performance of other methods deteriorated rapidly with an increasing number of outliers, LLM-Filter demonstrated minimal performance loss.

We also tested the classical observation-disturbance scenario \parencite{duran2024outlier}, where robust filtering methods such as EnKFI, EnKFS, and HubEnKF are known to excel. As shown in Table~\ref{tab.obs_corroput}, although LLM-Filter did not achieve the best results, it delivered performance comparable to specialized robust algorithms.
These findings highlight LLM-Filter's ability to generalize effectively even in the presence of significant outlier interference or system turbulence, offering a reliable solution across diverse and challenging scenarios.

In addition, we provide the complete experimental results for the cross-system scenarios discussed in Section~\ref{subsec.zero-shot_estimation}, as shown in Table~\ref{tab.zero-shot}.

\begin{table*}[ht!]
\centering
\caption{RMSE Results for Observation Outlier Scenarios.
Here \emph{$R$} denotes the true observation covariance with outliers, while \emph{$R_0$} represents the assumed covariance.}
\label{tab:results}
\resizebox{0.75\textwidth}{!}{
\begin{tabular}{@{}llcccccc@{}}
\toprule
{\textbf{Method}} & \textbf{$R$} & \textbf{LLM-Filter} & \textbf{EnKF} & \textbf{EnKFI} & \textbf{EnKFS} & \textbf{HubEnKF} \\ \midrule
\multirow{5}{*}{\textbf{Selkov}} 
    & $R_0$     & \textbf{0.4061}              & \underline{0.5978}        & 0.6703         & 0.6081         & 0.6676           \\
    & $25R_0$    & \textbf{0.6898}              & 1.8626        & 1.9874         & \underline{1.6523}         & 2.1998           \\
    & $50R_0$    & \textbf{0.7034}              & 2.5496        & 3.4248         & 2.6597         & \underline{2.5318}           \\
    & $75R_0$    & \textbf{0.7074}              & 3.0952        & 4.3829         & 3.1699         & \underline{2.8636}           \\
    & $100R_0$   & \textbf{0.7090}              & 3.5625        & 4.8417         & 4.3972         & \underline{2.9850}           \\ \midrule
\multirow{5}{*}{\textbf{Oscillator}}
    & $R_0$     & \textbf{0.5247}              & \underline{0.5505}        & 0.5516         & 0.5548         & 0.5540           \\
    & $25R_0$    & \textbf{0.5946}              & \underline{1.1630}        & 1.2714         & 1.1674         & 1.1746           \\
    & $50R_0$    & \textbf{0.6044}              & 1.3267        & \underline{1.3195}         & 1.3575         & 1.4778           \\
    & $75R_0$    & \textbf{0.6083}              & \underline{1.4701}        & 3.1428         & 1.4757         & 1.8707           \\
    & $100R_0$   & \textbf{0.6110}              & 1.6154        & 1.6040         & \underline{1.5793}         & 2.1680           \\ \midrule
\multirow{5}{*}{\textbf{Hopf}} 
    & $R_0$     & \textbf{0.5751}              & \underline{0.6322}        & 0.6983         & 0.6380         & 0.6982           \\
    & $25R_0$    & \textbf{0.7142}              & 0.9946        & 1.1299         & \underline{0.9938}         & 1.0976           \\
    & $50R_0$    & \textbf{0.7738}              & \underline{1.0549}        & 1.1657         & 1.2035         & 1.1155           \\
    & $75R_0$    & \textbf{0.8010}              & \underline{1.0898}        & 1.1872         & 1.5071         & 1.1313           \\
    & $100R_0$   & \textbf{0.8096}              & \underline{1.1146}        & 1.1989         & 2.1321         & 1.1483           \\ \bottomrule
\end{tabular}
}
\label{tab.cross_outliers}
\end{table*}

\begin{table*}[ht!]
\caption{RMSE Results of Observation-Disturbance Scenarios.}
\centering
\resizebox{0.65\textwidth}{!}{\begin{tabular}{lccccc}
\toprule
\multicolumn{1}{c}{\textbf{Method}} & \textbf{LLM-Filter} &  \textbf{EnKF} & \textbf{EnKFI} & \textbf{EnKFS} & \textbf{HubEnKF}  \\
\midrule
\multicolumn{1}{c}{\textbf{Selkov}} & \textbf{0.4999}  & 0.9223 & 0.9500 & \underline{0.6079} & 0.6198  \\
\multicolumn{1}{c}{\textbf{Oscillator}} & 0.7451  & 1.3431 & 1.3611 & \underline{0.5543} & \textbf{0.5364} \\
\multicolumn{1}{c}{\textbf{Hopf}} & 0.6827  & 0.8220 & NaN & \underline{0.6379} & 0.7226  \\
\multicolumn{1}{c}{\textbf{Lorenz96}} & \textbf{0.9186} & 7.6225 & 7.7459 & 4.5833 & \underline{3.1399}  \\
\multicolumn{1}{c}{\textbf{VL20}} & \textbf{0.8876}  & NaN & 7.7777 & 11.5041 & \underline{0.9684}  \\
\bottomrule
\end{tabular}}
\label{tab.obs_corroput}
\end{table*}

\begin{table*}[ht!]
\centering
\caption{RMSE of Cross-System Scenarios.}
\resizebox{0.85\textwidth}{!}{
\begin{tabular}{lccccc}
\toprule
\multicolumn{1}{c}{\textbf{Method}} & \textbf{LLM-Filter} & \textbf{LLM-Filter-O} & \textbf{MEstimator} & \textbf{RStateNet} & \textbf{ProTran}  \\ \midrule
\multicolumn{1}{c}{\textbf{Tracking$\rightarrow$Pendulum}} & \textbf{1.2568} &\underline{1.6573}& 3.6957 & 3.1052 & 3.1162  \\
\multicolumn{1}{c}{\textbf{Pendulum$\rightarrow$Tracking}} & \textbf{1.4214} &\underline{1.8936}& 2.7407 & 5.0927 & NaN  \\
\midrule
\multicolumn{1}{c}{\textbf{Oscillator$\rightarrow$Hopf}} & \textbf{0.6613} &\underline{2.1125}& NaN & 3.8645 & 3.5668  \\
\multicolumn{1}{c}{\textbf{Hopf$\rightarrow$Oscillator}} & \textbf{0.9069}&\underline{1.5624} & 2.5369 & 2.7542 & 3.0075  \\
\bottomrule
\label{tab.zero-shot}
\end{tabular}}
\end{table*}

\subsection{Scaling Behavior}
We investigate the scaling behavior of LLM-Filter in state estimation tasks by evaluating models with various pretrained LLM backbones. Detailed results can be found in Table~\ref{tab.various_llms}. 
Our analysis reveals that LLM-Filter follows scaling-law behavior: as the parameter size of the model increases, the average estimation error tends to decrease, leading to improved estimation performance.

\begin{table*}[ht!]
\centering
\caption{RMSE Results of Various LLMs in the LLM-Filter Framework.}
\resizebox{0.7\textwidth}{!}{
\begin{tabular}{lccccc}
\toprule
\multicolumn{1}{c}{\textbf{LLM}} & \textbf{GPT-2 (124M)} & \textbf{OPT-1.3B} & \textbf{LLaMA-7B} & \textbf{LLaMA-13B} \\
\midrule
\multicolumn{1}{c}{\textbf{Selkov}} & 1.0434 & 1.1713 & \underline{0.6369} & \textbf{0.6200} \\
\multicolumn{1}{c}{\textbf{Oscillator}} & 0.9393 & 1.0702 & \underline{0.5753} & \textbf{0.4441} \\
\multicolumn{1}{c}{\textbf{Hopf}} & 0.9611 & 0.8466 & \underline{0.8180} & \textbf{0.7000} \\
\multicolumn{1}{c}{\textbf{Lorenz96}} & 0.9727 & \underline{0.9029} & 0.9735 & \textbf{0.9147} \\
\multicolumn{1}{c}{\textbf{VL20}} & 0.9931 & \textbf{0.5999} & 0.8433 & \underline{0.7623} \\
\midrule
\multicolumn{1}{c}{\textbf{Average}} & 0.9819 & 0.9182 & \underline{0.7694} & \textbf{0.6882} \\
\bottomrule
\end{tabular}}
\label{tab.various_llms}
\end{table*}

\subsection{Hyperparameter Sensitivity}
\label{subsec.hyberparameter sensitivity}
\begin{figure*}[ht]
\begin{center}
\centerline{\includegraphics[width=0.99\textwidth]{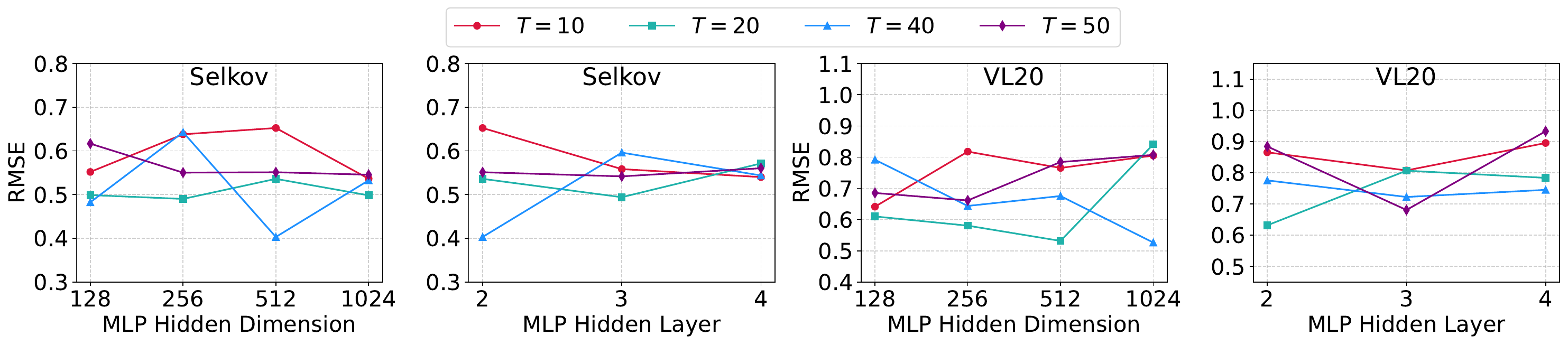}}
\caption{Hyperparameter Sensitivity of LLM-Filter on the Selkov and VL20 Systems.}
\label{fig.sensitive}
\end{center}
\vspace{-10pt}
\end{figure*}

In the main experiments, the configuration of the LLM-Filter is provided in Appendix \ref{apd.implementation}.
Here, we verify the robustness of the LLM-Filter with respect to hyperparameters, including the sliding length $T$, the hidden layer dimension, and the layer number of MLP in $\operatorname{ObsEmbedding}$ and $\operatorname{StateProjection}$. 
Experiments were conducted on the Selkov and VL20 systems, with the results summarized in Figure~\ref{fig.sensitive}.
Overall, LLM-Filter demonstrates consistent performance across different parameter settings, showing it is largely insensitive to hyperparameter variations. On average, a window length of 20 or 40 yields better results. 
This suggests that appropriately chosen window lengths strike a balance—shorter windows risk losing crucial information, while longer windows may fail to capture local features effectively.

Additionally, the segment length in LLM-Filter is a hyperparameter that determines how many observations are grouped into a single token before being fed into the LLM.
We conducted further experiments using different segment lengths while keeping the window length fixed at 40, as shown in Table~\ref{tab:context_length}.
The performance of LLM-Filter remains relatively stable across different segment lengths, demonstrating its robustness to this hyperparameter.
\clearpage

\begin{table}[htbp]
\centering
\caption{RMSE Comparison with Varying Segment Lengths}
\resizebox{0.5\textwidth}{!}{
\begin{tabular}{lccc}
\toprule
\textbf{Segment Length} & \textbf{Base (20)} & \textbf{10} & \textbf{5} \\
\midrule
Selkov     & 0.4061 & 0.7024 & 0.3077 \\
Oscillator & 0.5247 & 0.5829 & 0.5298 \\
Hopf       & 0.5751 & 0.6222 & 0.5779 \\
Pendulum   & 0.8348 & 0.5028 & 0.8254 \\
Lorenz96   & 0.9149 & 0.9200 & 0.9182 \\
VL20       & 0.7717 & 1.0346 & 1.0221 \\
\midrule
\textbf{Average} & \textbf{0.6712} & \textbf{0.7275} & \textbf{0.6969} \\
\bottomrule
\end{tabular}}
\label{tab:context_length}
\end{table}

\end{document}